\documentclass[10pt, conference, compsocconf]{IEEEtran}

\usepackage{ifpdf}
\usepackage{cite}
\ifCLASSINFOpdf
\else
\fi
\usepackage[cmex10]{amsmath}
\usepackage{algorithmic}
\usepackage{array}
\usepackage{mdwmath}
\usepackage{mdwtab}
\usepackage{eqparbox}
\usepackage{fixltx2e}
\usepackage{stfloats}
\usepackage{url}
\usepackage{amssymb,subcaption,adjustbox}
\usepackage{multirow,color}

\newcommand{\etal}{\textit{et al}.}
\newcommand{\ie}{\textit{i}.\textit{e}., }

\begin{document}
%\setlength\baselineskip{\fill}
%\setlength\lineskip{\fill}
%\setlength\parskip{\fill}
%
% paper title
% can use linebreaks \\ within to get better formatting as desired
\title{Total-Text: A Comprehensive Dataset for Scene Text Detection and Recognition \\ (EXTENDED VERSION)}

% author names and affiliations
% use a multiple column layout for up to two different
% affiliations

\author{\IEEEauthorblockN{Chee Kheng Ch'ng \,\,\,\,\,\,\, Chee Seng Chan}
\IEEEauthorblockA{Centre of Image \& Signal Processing, Faculty of Computer Science \& Info. Technology,\\
University of Malaya, Malaysia\\
chngcheekheng@siswa.um.edu.my, cs.chan@um.edu.my}
}

% make the title area
\maketitle

\begin{abstract}
Text in curve orientation, despite being one of the common text orientations in real world environment, has close to zero existence in well received scene text datasets such as ICDAR'13 and MSRA-TD500. The main motivation of Total-Text is to fill this gap and facilitate a new research direction for the scene text community. On top of conventional horizontal and multi-oriented text, it features curved-oriented text. Total-Text is highly diversified in orientations, more than half of its images have a combination of more than two orientations. Recently, a new breed of solutions that casted text detection as a segmentation problem has demonstrated their effectiveness against multi-oriented text. In order to evaluate its robustness against curved text, we fine-tuned DeconvNet and benchmark it on Total-Text. Total-Text with its annotation is available at https://github.com/cs-chan/Total-Text-Dataset.
 \end{abstract}


\IEEEpeerreviewmaketitle

\section{Introduction}
Scene text detection is one of the active computer vision topics due to the growing demands of applications such as multimedia retrieval, industrial automation, assisting device for vision-impaired people, etc. Given a natural scene image, the goal of text detection is to determine the existence of text, and return the location if it is present. 

Well known public datasets such as ICDAR'03, '11, '13\cite{karatzas2013icdar} (term as ICDARs from here onwards), and MSRA-TD500 \cite{yao2012detecting} have played a significance role in initiating the momentum of scene text related research. One similarity in all the images of ICDARs is that all the texts are in horizontal orientation\cite{survey}. Such observation has inspired researchers to incorporate horizontal assumption \cite{zhang2015symmetry,huang2013text,neumann2013scene,huang2014robust,xiangcvpr2017} in solving the scene text detection problem. In 2012, Yao \etal~\cite{yao2012detecting} introduced a new scene text dataset, namely MSRA-TD500, that challenged the community with texts arranged in multiple orientations. The popularity of it in turn defined the convention of `multi-oriented' texts. However, a closer look into the MSRA-TD500 dataset revealed that most, if not all the texts are still arranged in a straight line manner as to ICDARs (more details in Section \ref{sect3}). Curved-oriented texts(term as curved text from here onwards), despite its commonness, are missing from the context of study. To the best of our knowledge, CUTE80 \cite{risnumawan2014robust} is the only available scene text dataset to-date with curved text. However, its scale is too small with only 80 images and it has very minimal scene diversity.

\begin{figure}[t]
	\begin{center}
		\includegraphics[height=0.8\linewidth, width=\linewidth]{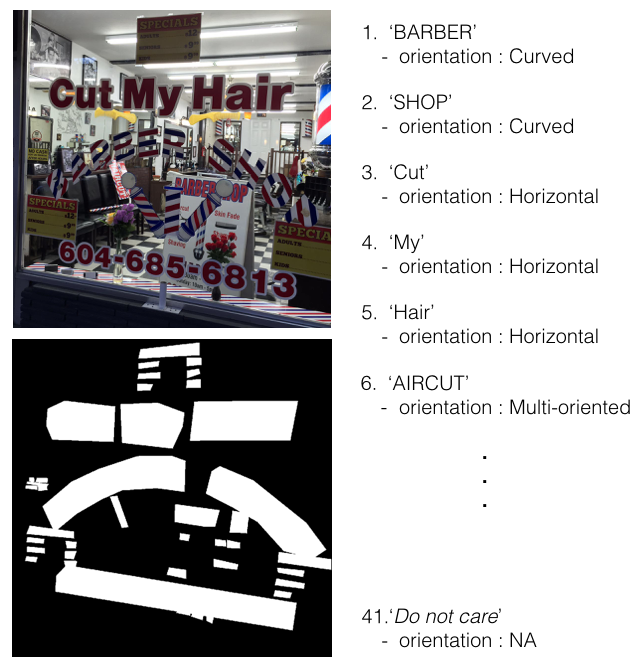}
	\end{center}
	\caption{Annotation details of Total-Text, including transcription, polygon-shaped and rectangular bounding box vertices, orientations, \textit{care} and \textit{do not care} regions, and binary mask.} \vspace{-.1in}
	\label{fig:cover}
\end{figure}

\begin{figure*}[ht]
	\begin{center}
		\includegraphics[height=0.14\linewidth, width=0.95\linewidth]{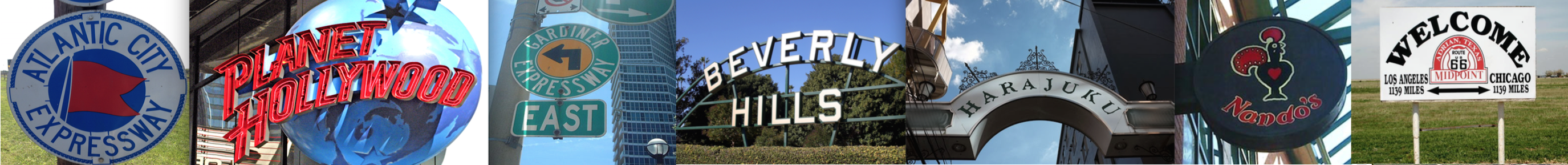}	
	\end{center}
	\caption{Curved text is commonly seen in real world scenery.} \vspace{-.1in}
	\label{fig:t1a}
\end{figure*} 

Without the motivation of a proper dataset, effort in solving the curved text detection problem is rarely seen. This phenomenon brings us to our primary contribution of this paper: Total-Text, a scene text dataset collected with curved text in mind, filling the gap in scene text datasets in terms of text orientations. It has 1,555 scene images, 9,330 annotated words with 3 different text orientations including horizontal, multi-oriented, and curved text. 

Orientation assumption is commonly seen in text detection algorithms. We believe that the heuristic design to cater different types of text orientations hold back the generalization of text detecting system against texts in the real world with unconstrained orientations. Recent works \cite{Zhang_2016_CVPR,he2016accurate,yao2016scene} have started to cast text detection as a semantic segmentation problem, and achieved state-of-the-art results in ICDAR'11, '13 and MSRA-TD500 datasets. They have reported successful detection of curved text as well. He et al.[3] system in particular has no orientation assumption and hueristic grouping mechanism. This bring us to the secondary contribution of this paper, we looked into this new solution and revealed how it handle multiple oriented text in natural scene.

\section{Related Works}
\label{related}

This section will discuss closely related works, specifically scene text datasets and text detection system. For completeness, readers are recommended to read \cite{survey}. 

\begin{figure*}[ht]
	\begin{center}
		\includegraphics[height=0.25\linewidth, width=0.9\linewidth]{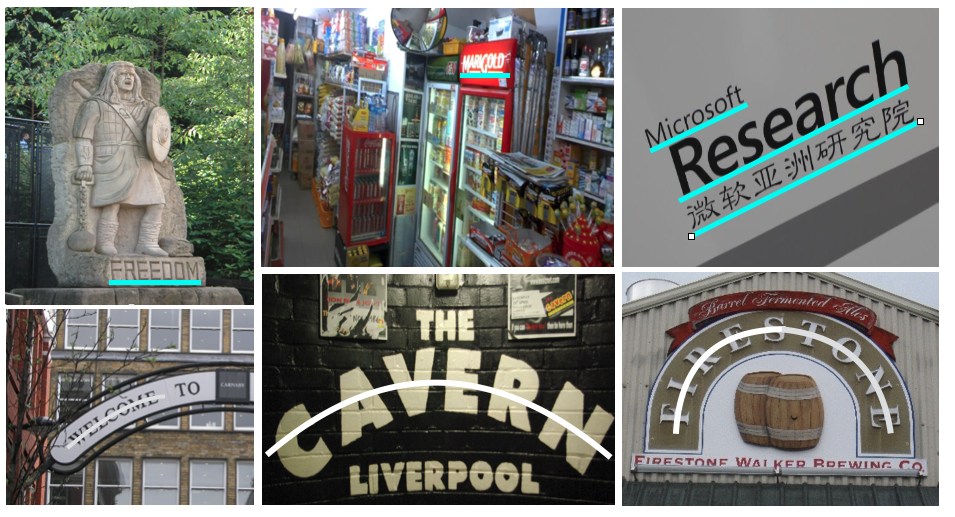}
	\end{center}
	\caption{1st row: Examples from ICDAR 2013, ICDAR2015 and MSRA-TD500; 2nd row: Slightly curved to extremely curved text examples from the Total-Text.}
	\label{fig:def}\vspace{-.1in}
\end{figure*}
\begin{figure*}[t]
	\begin{subfigure}{0.7\linewidth}
		\centering
		\includegraphics[height=0.2\linewidth, width=0.95\linewidth]{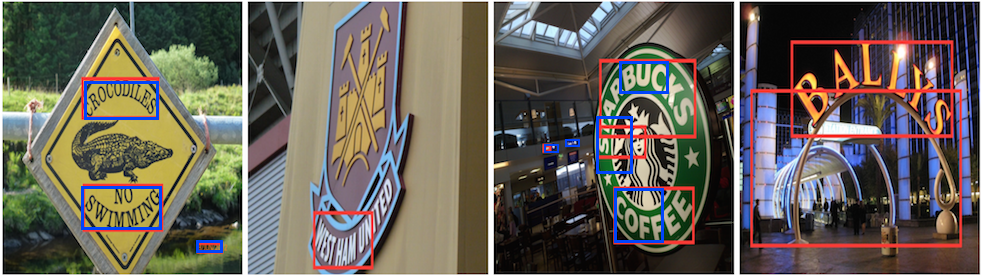}
		\caption{Yin et. al. \cite{yin2013robust} (red bounding box) and Huang \etal \cite{huang2014robust} (blue bounding box)}
		\label{result1a}
	\end{subfigure}
	\begin{subfigure}{0.25\linewidth}
		\centering
		\includegraphics[height=0.56\linewidth, width=0.9\linewidth]{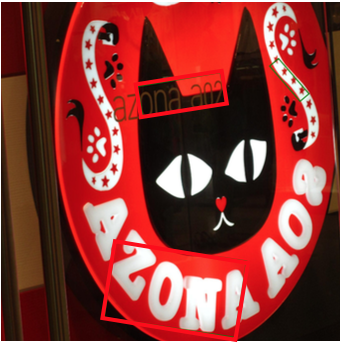}
		\caption{Shi \etal \cite{xiangcvpr2017}}
		\label{result1b}
	\end{subfigure}
	\caption{These show that the current state-of-the-art solutions could not detect curved text effectively.} 
	\label{fig:result1}
\end{figure*}

\subsection{Scene Text Datasets}

\textbf{ICDARs}\cite{karatzas2013icdar} has three variants. ICDAR'03 started out with 509 camera taken scene text images. All the scene texts in the dataset appear in horizontal orientation. In ICDAR'11, the total number of images were reduced to 484 to eliminate duplication in the previous version. ICDAR'13 further trimmed down the 2011 version to 462 images in total. Improvement was done to increase its text categories and tasks. In ICDAR'13, there are 462 images of horizontal English texts. Recently, ICDAR launched a new challenge \cite{karatzas2015icdar} named as the `Incidental Scene Text' (also known as the ICDAR'15), which is based on 1670 images captured with wearable devices. It is more challenging than previous datasets as it has included text with arbitrary orientation and most of them are out of focus.

\textbf{MSRA-TD500}\cite{yao2012detecting} was introduced in 2012 to address the lack of arbitrary orientated text in scene text datasets. It has 300 training and 200 testing images; annotated with minimum area rectangle.

\textbf{COCO-text}\cite{veit2016cocotext} was released in the early 2016, and is the largest scene text dataset to-date with 63,686 images and 173,589 labeled text regions. This large scale dataset contains all variety of text orientations: horizontal, arbitrary and curved. However, it used the axis oriented rectangle as groundtruth, which seems to be applicable only to horizontal and vertical texts. 

\textbf{CUTE80}\cite{risnumawan2014robust} is the only curved text dataset available in public to the best of our knowledge. It has only 80 images and limited sceneries.

%------------------------------------------------------------------
\subsection{Scene Text Detection:} 

Scene text detection has seen significant progress after the seminal work by Epshtein \etal \cite{epshtein2010detecting} and Neumann and Matas \cite{matas2004robust}. In the former, Stroke Width Transform (SWT) was proposed to detect text. This method considered similar stroke widths to group text components and studied the component properties to classify them. In the latter, Maximally Stable Extremal Regions (MSER) was exploited to extract text components. They used geometrical properties of the components and a classifier to detect text. Both represent character better than all other feature extractors like color, edge, texture and etc. Upon picking up potential character candidates, these connected components based algorithms typically go through text line generation, candidates filtering and segmentation as pointed out by this survey \cite{survey}. 

As to many other computer vision tasks, the incorporation of Convolutional Neural Network (CNN) in localizing text is a very active research at the moment. Huang \etal~\cite{huang2014robust} trained a character classifier to examine components generated by MSER, with the objective of improving the robustness of feature extraction process. Alongside this work, \cite{wang2012end,jaderberg2014deep} also trained a CNN to classify text components from non-text. This line of work demonstrated the high discriminative power of CNN as a feature extractor. However, interestingly, Zhang \etal~\cite{Zhang_2016_CVPR} argued that leveraging on CNN as a character detector has restricted the CNN's potential due to the local nature of characters. Zhang \etal~trained two Fully Convolutional Networks (FCN) \cite{long2015fully}: 1) A Text-Block FCN that considers both local and global contextual info at the same time to identify text regions in an image, 2) Character-Centroid FCN to eliminate false text line candidates. However, text line generation, which plays a key role in grouping characters into a word, did not receive much benefit from the robust CNN. While most of the algorithms \cite{Zhang_2016_CVPR,jaderberg2014deep} handcrafted the text line generation process, He \etal~\cite{he2016accurate} trained a FCN to infer text line candidates. By cascading a text region and a text line using supervised FCN, Cascaded Convolution Text Network (CCTN) achieved generalization in terms of text orientations, and is one of the best performing system in both horizontal and abritrary oriented scene text datasets: ICDAR 2013 and MSRA-TD500. %Recently, Shi \etal \cite{xiangcvpr2017} presented SegLink, an oriented text detection method and achieved the best result in ICDAR 2015 Incident dataset.
%----------------------------------------------------------------------------------------------
\section{Total-Text Dataset}
\label{sect3}
This section will discuss a) the motivation of collecting Total-Text; b) observation made on horizontal, multi-oriented, and curved text; c) orientation assumption aspect in the current state-of-the-art algorithms, and d) different aspects and statistics of Total-Text.

\subsection{Dataset Attributes}
\begin{figure*}[ht]
	\centering
	\begin{subfigure}{0.23\linewidth}
		\centering
		\includegraphics[height=0.9\textwidth, width=0.9\textwidth]{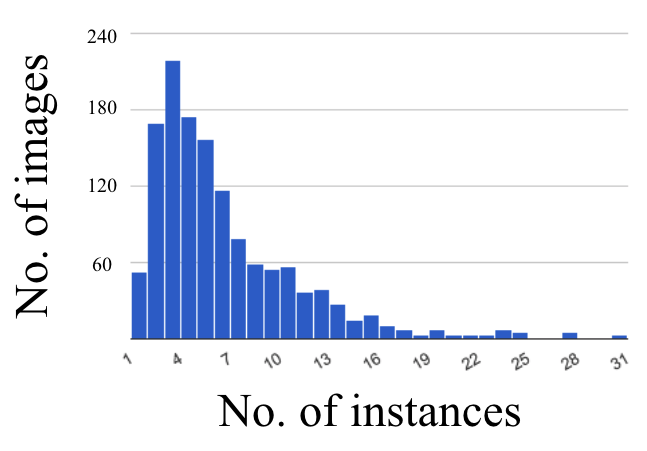}
		\caption{Text instances per image}
		\label{unreal31}
	\end{subfigure}
	\begin{subfigure}{0.23\linewidth}
		\centering
		\includegraphics[height=0.9\textwidth, width=0.9\textwidth]{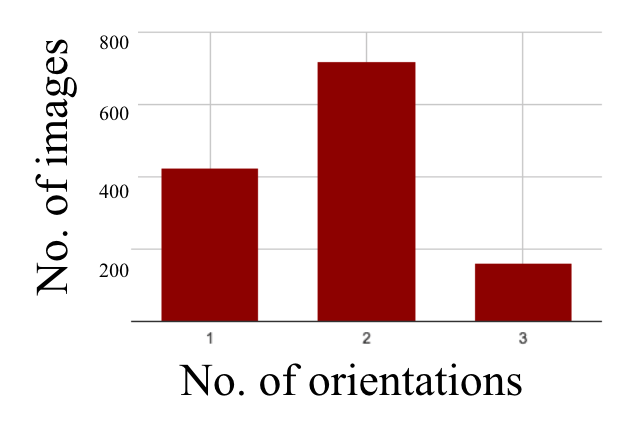}
		\caption{Text orientations per image}
		\label{unreal21}
	\end{subfigure}
	\begin{subfigure}{0.23\linewidth}
		\centering
		\includegraphics[height=0.9\textwidth, width=0.9\textwidth]{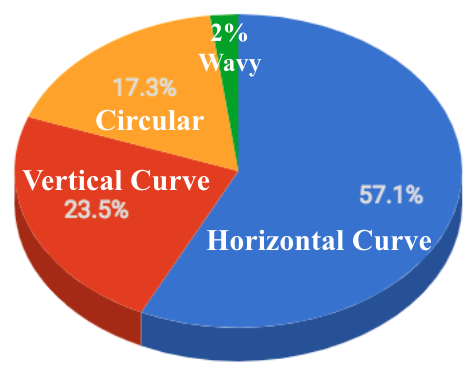}
		\caption{Curve variations}
		\label{unreal11}
	\end{subfigure}
	\begin{subfigure}{0.23\linewidth}
		\centering
		\includegraphics[height=0.9\textwidth, width=1\textwidth]{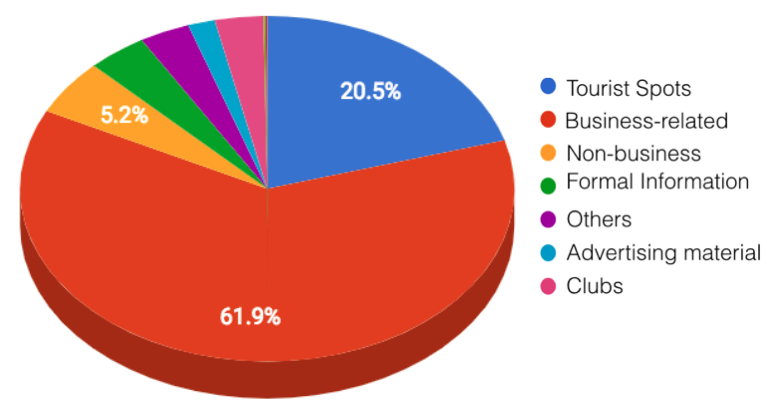}
		\caption{Occurrence of curved text}
		\label{unreal41}
	\end{subfigure}
	\caption{Statistics of Total-Text dataset}
	\label{fig:stats}
\end{figure*}

\textbf{Curved text is an overlooked problem.} 
The effort of collecting this dataset is motivated by the missing of curved text in existing scene text datasets. Curved text can be easily found in real life scenes such as: business logos, signs, entrances etc as depicted in Fig. \ref{unreal41}, surprisingly such data has close to zero existence in the current datasets \cite{karatzas2013icdar,karatzas2015icdar,yao2012detecting}. The most popular scene text dataset over the decade, ICDARs have only horizontal text \cite{survey}. Consequently, vast majority of algorithms assume text linearity to tackle the problem effectively. As a result of overwhelming attention, performances of text detections in ICDARs are saturated at quite a high point (0.9 in terms of f-score). Meanwhile, multi-oriented text also received a certain amount of attention from this community. MSRA-TD500 is a well known dataset that introduced this challenge to the field. Algorithms like \cite{Zhang_2016_CVPR,yin2015multi} were designed to cater multi-oriented text. To the best of our knowledge, scene text detection algorithms designed for curved orientation \cite{risnumawan2014robust} in consideration is relatively unpopular. We believe that the lack of such dataset is the obvious reason why the community has overlooked it. Hence, we propose Total-Text with 4,265 curved text out of 9,330 total text instances, hoping to spur an interest in the community to address curved text.

\textbf{Curved text observation.} Geometrically speaking, a straight line has no angle variation along the line, and thus can be described as a linear function, $y = mx + c$. {\it A curved line is not a straight line}. It is free of angle variation restriction throughout the line. Shifting to the scene text perspective, we observed that horizontal oriented text or word is a series of characters that can be connected by a straight line; their bottom alignment in particular for most cases. At the same time, multi-oriented text, in scene text convention, can also be connected by a straight line, given an offset with respect to a horizontal line. Meanwhile, characters a in curved word will not have unified angle offset, in which deemed to fit a polynomial line in text level (refer to Fig. \ref{fig:def} for image examples). In our dataset collection, we found out that curved text in natural images could vary from slightly curved to extremely curved. Also, it is not surprising to find that most of them are in the shape of a symmetric arc due to the symmetrical preferences in human vision \cite{adams2011science}. 

\begin{figure*}[ht]
	\begin{center}
		\begin{subfigure}{0.23\textwidth}{{\includegraphics[height=0.65\linewidth, width=0.9\linewidth]{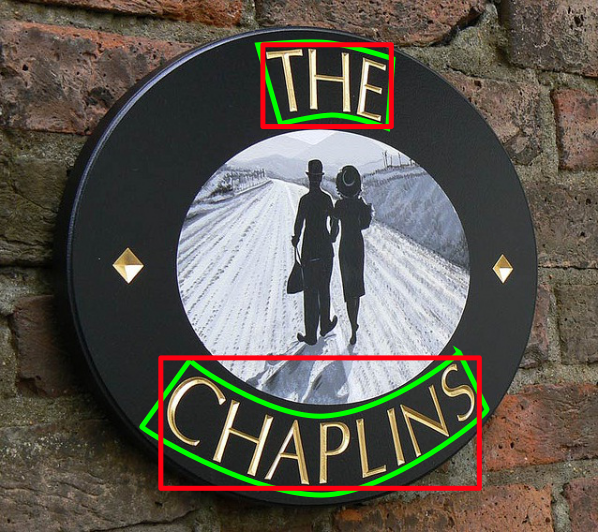}}}\end{subfigure} 
		\begin{subfigure}{0.23\textwidth}{{\includegraphics[height=0.65\linewidth, width=0.9\linewidth]{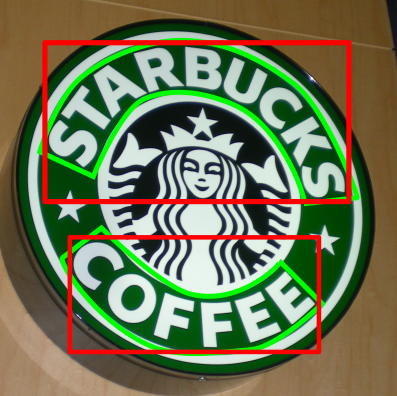}}}\end{subfigure}
		\begin{subfigure}{0.23\textwidth}{{\includegraphics[height=0.65\linewidth, width=0.9\linewidth]{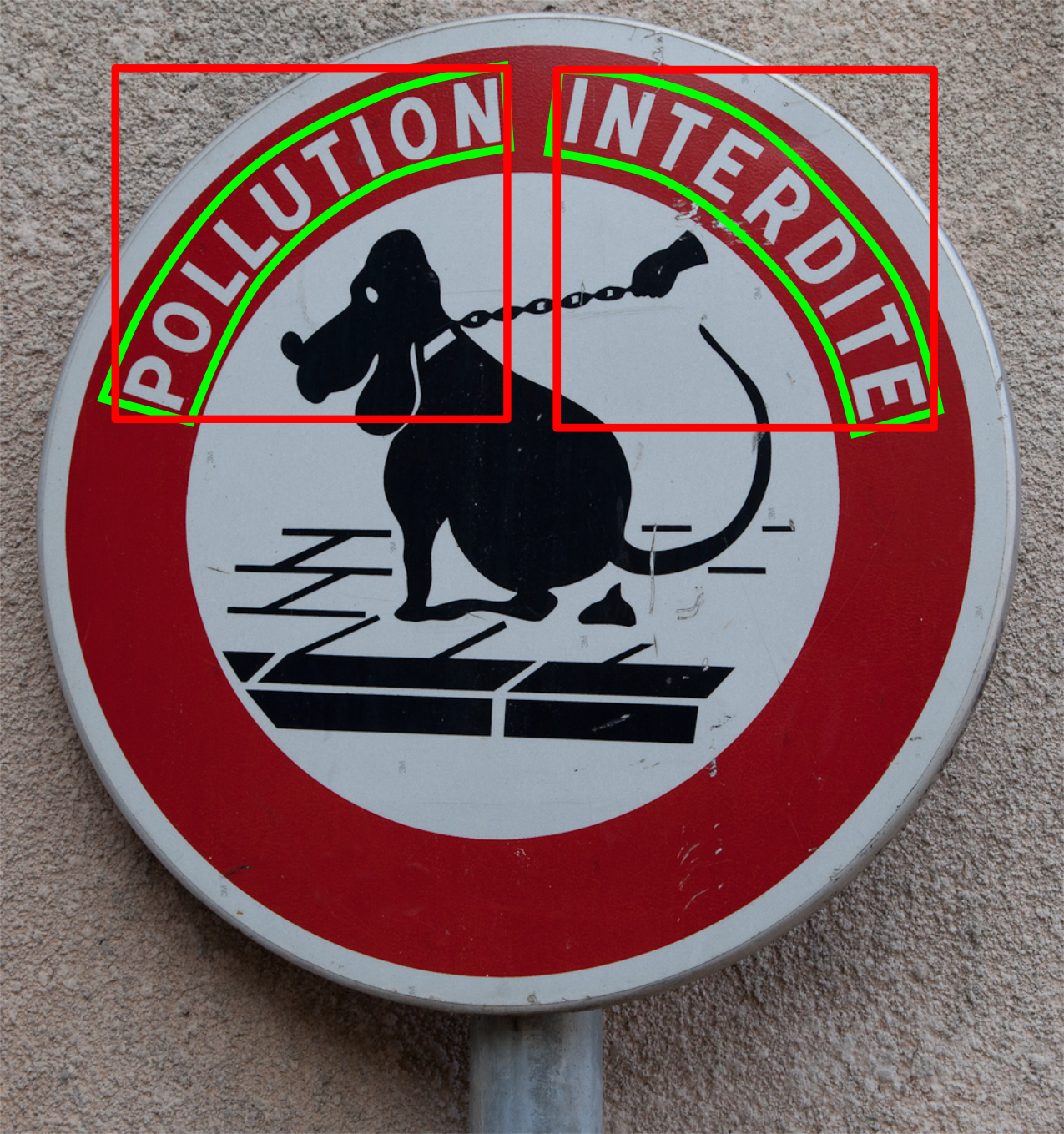}}}\end{subfigure}
		\begin{subfigure}{0.23\textwidth}{{\includegraphics[height=0.65\linewidth, width=0.9\linewidth]{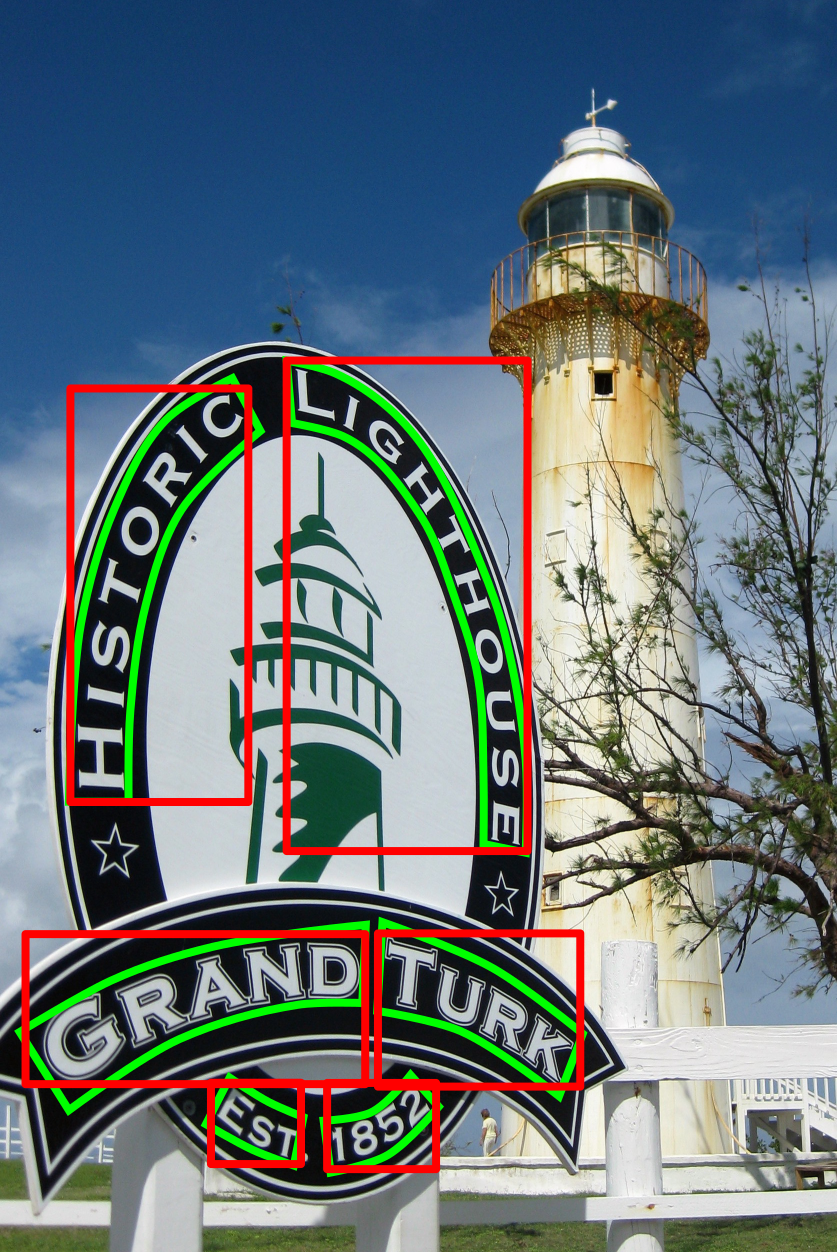}}}\end{subfigure}
	\end{center}
	\caption{Comparison between conventional rectangular bounding box (red colour) and the proposed polygon-shaped bounding region (green colour) in Total-Text. Polygon-shaped appeared to be the better candidate for groundtruth.}\vspace{-.1in}
	\label{fig:diffgt}
\end{figure*}
\begin{figure*}[ht]
	\centering
	\begin{subfigure}{0.5\linewidth}
		\centering
		\captionsetup{justification=centering}
		\includegraphics[height=0.25\linewidth, width=0.95\linewidth]{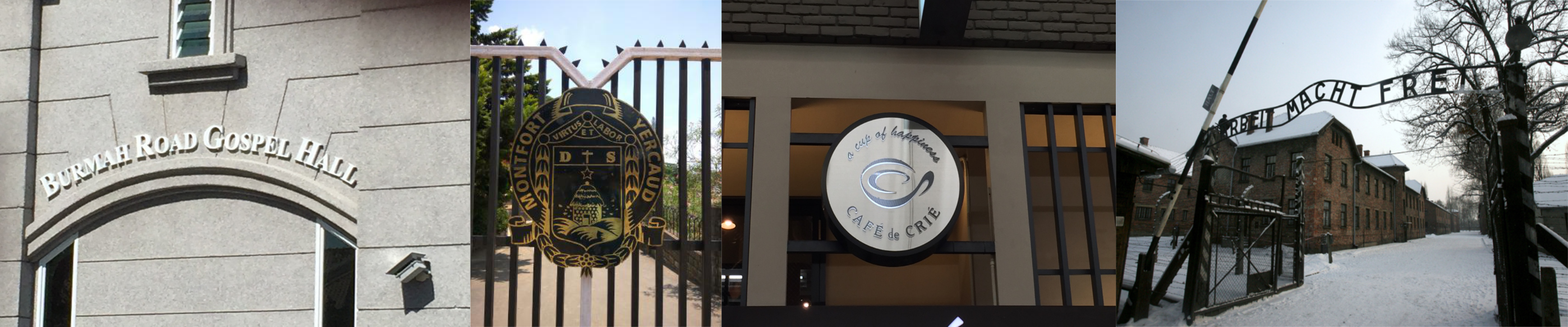}
		\includegraphics[height=0.25\linewidth, width=0.95\linewidth]{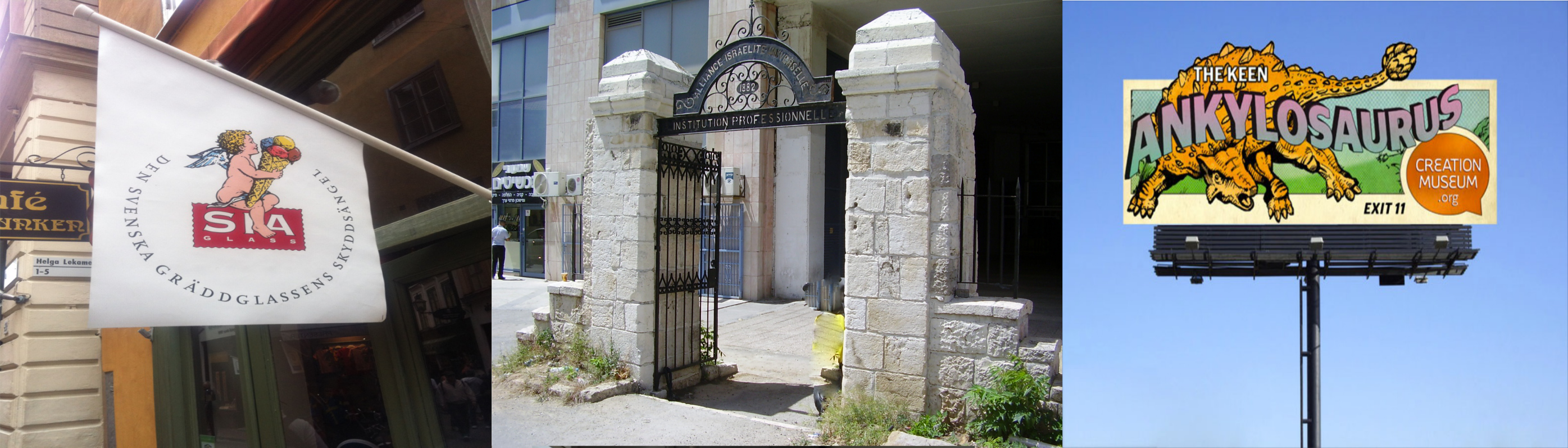}
		\includegraphics[height=0.26\linewidth, width=0.95\linewidth]{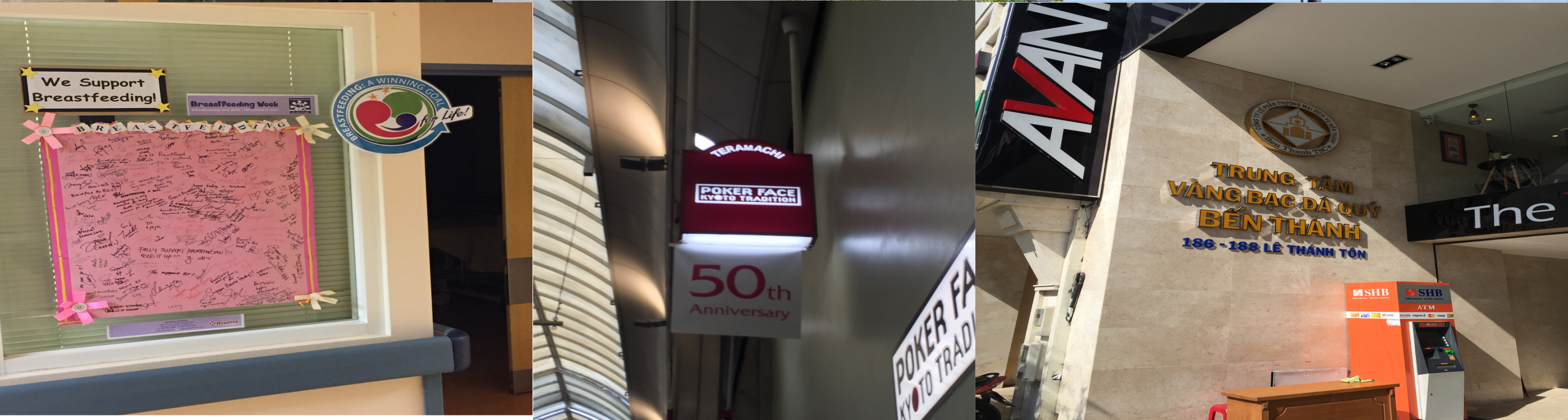}
		\subcaption{Various text orientations (from left to right).\\
			\textbf{Top} (One orientation): HC; VC; Cir and W.\\ \textbf{Middle} (Two orientations): Cir+H; MO+HC; W+H.\\ \textbf{Bottom} (Three orientations): H+MO+VC; H+MO+HC; H+MO+Cir}
		\label{fig:examples_a}
	\end{subfigure}
	\hfill
	\begin{subfigure}{0.49\linewidth}
		\centering
		\includegraphics[height=0.7\linewidth, width=0.98\linewidth]{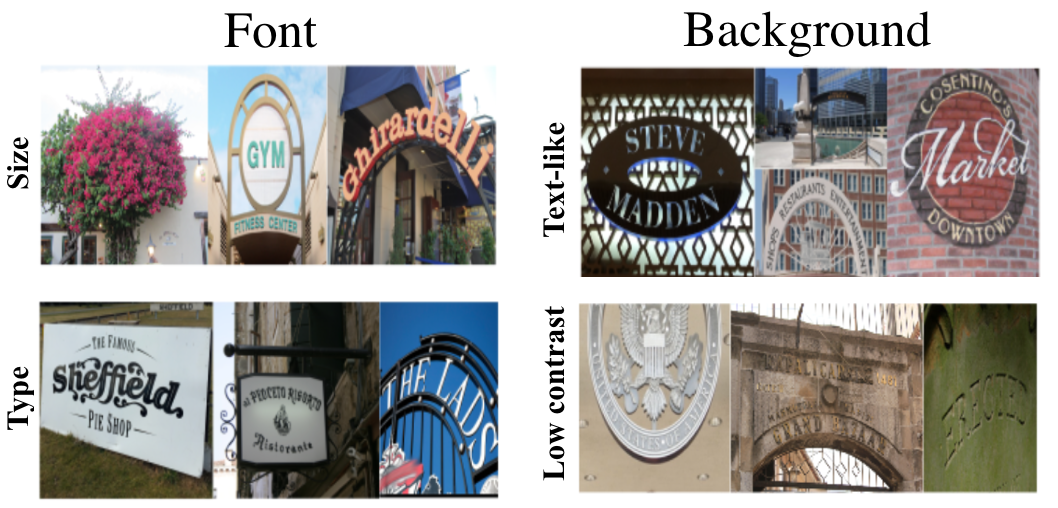}
		\caption{Various text fonts and image backgrounds}
		\label{fig:examples_b}
	\end{subfigure}
	\caption{Total-Text dataset is challenging due to its highly diversified orientation compositions and scenery.\\ Legends: H=horizontal, MO=multi-oriented, HC=horizontal curve, VC=vertical curve, Cir=circular and W=Wavy.}\vspace{-.1in}
	\label{fig:examples}
\end{figure*}

\textbf{Orientation assumption.} We observed that orientation assumption is a must in a lot of algorithms \cite{zhang2015symmetry,huang2013text,neumann2013scene,huang2014robust,Zhang_2016_CVPR,yin2015multi}. We took a closer look into the orientation assumption aspect of existing text detection algorithms and see how it fits into the observation we have made on the curved text. We mainly focused on systems in which the authors claimed to have multi-oriented text detection capability and reported their results on MSRA-TD500. Zhang \etal~\cite{Zhang_2016_CVPR} first used the FCN to create a saliency map and generate text blocks. Consequently, the system draw a straight line from the middle point of the generated text blocks, aiming to hit as many character components as possible; the straight line with the angle offset that hit the most text blocks will be considered as text line for the subsequent step. We believe that such mechanism would not work in our dataset, as a straight line would miss the polynomial nature of curved text. \cite{yin2015multi} focused on the text candidate construction part to detect multi-oriented text. Their algorithm will first clusters character pairs with consistent orientation or perspective view into the same group. As we can see in Fig. \ref{fig:def} (second row, second and third image specifically), characters in a single curved word could have multiple variations in terms of orientation. In fact, both of these algorithms, along with \cite{xiangcvpr2017}, have reported their failure on the same curved text images in MSRA-TD500 as illustrated in Fig. \ref{result1b}. It is worth to note that MSRA-TD500 has only 2 curved text instances in the entire dataset. Last but not least, we ran \cite{yin2013robust} and \cite{huang2014robust} on several images of Total-Text, results can be seen in Fig. \ref{fig:result1}.

\textbf{Focused scene text as a start.} Two of the latest scene text datasets, COCO-text and ICDAR 2015 emerged to challenge current algorithms with incidental images. For example, scene images in the ICDAR 2015 \cite{karatzas2015icdar} were captured without prior effort in positioning the text in it. Although it was not mentioned explicitly, one can deduce the emergence of these datasets are possibly due to: i) Performances of various algorithms on previous ICDARs dataset have saturated at a rather high point, hence a new dataset with higher level of complexity is deem required, ii) Well focused scene text are not likely to be captured by devices in real world scenarios. While the work done in curved text detection is considerably rare, we believe that it is at its infant stage. Inspired by the improvement in scene text detection and recognition brought by focused scene text datasets, notably ICDARs, and MSRA-TD500, we believe that focused scene text instead of incidental scene text is more appropriate to kick start related research work.

\textbf{Tighter groundtruth is better.} ICDAR 2015 employed quadrilaterals in its annotation to cater perspective distorted text \cite{karatzas2015icdar}. However, COCO-text used rectangular bounding boxes \cite{veit2016cocotext} like ICDAR 2013, which we think is a poor choice considering the text orientation variations in it. Fig. \ref{fig:diffgt} illustrates the downside of such bounding box annotation. Text regions cover much of the background which is not an ideal groundtruth for both evaluation and training. In Total-Text, we annotated the text region with polygon shapes that fits tightly, and the groundtruth is provided in polygon vertices format. 

\begin{figure*}[ht]
	\begin{center}
		\begin{subfigure}{0.23\textwidth}{{\includegraphics[height=0.34\linewidth, width=0.9\linewidth]{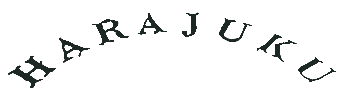}}}\end{subfigure}
		\begin{subfigure}{0.23\textwidth}{{\includegraphics[height=0.34\linewidth, width=0.9\linewidth]{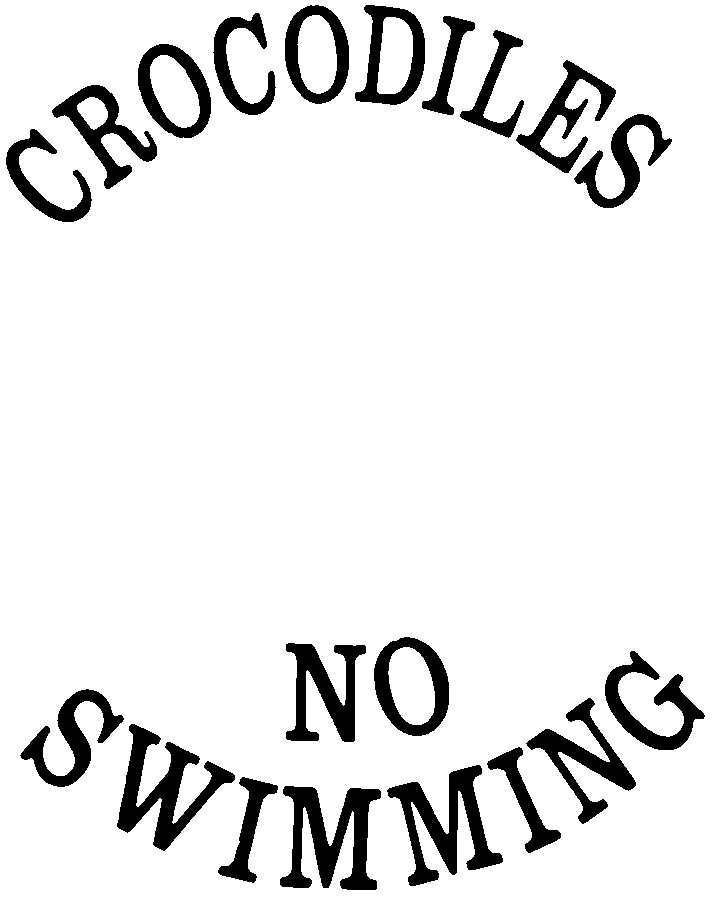}}}\end{subfigure}
		\begin{subfigure}{0.23\textwidth}{{\includegraphics[height=0.34\linewidth, width=0.9\linewidth]{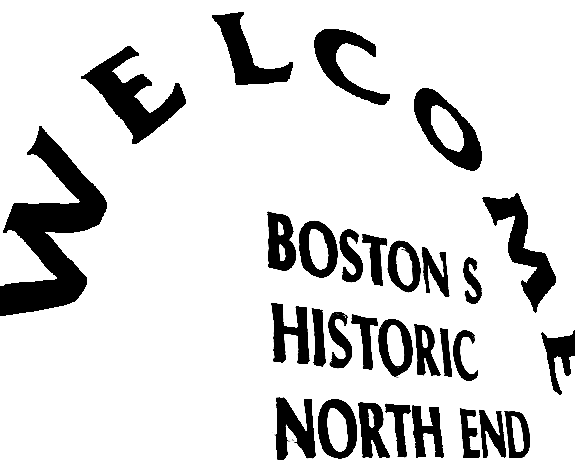}}}\end{subfigure}
		\begin{subfigure}{0.23\textwidth}{{\includegraphics[height=0.34\linewidth, width=0.9\linewidth]{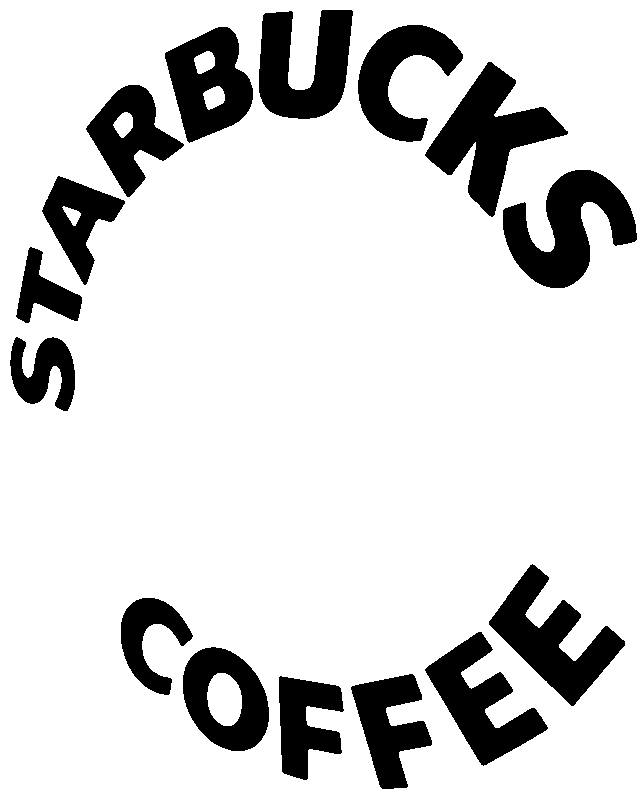}}}\end{subfigure}
		\begin{subfigure}{0.23\textwidth}{{\includegraphics[height=0.34\linewidth, width=0.9\linewidth]{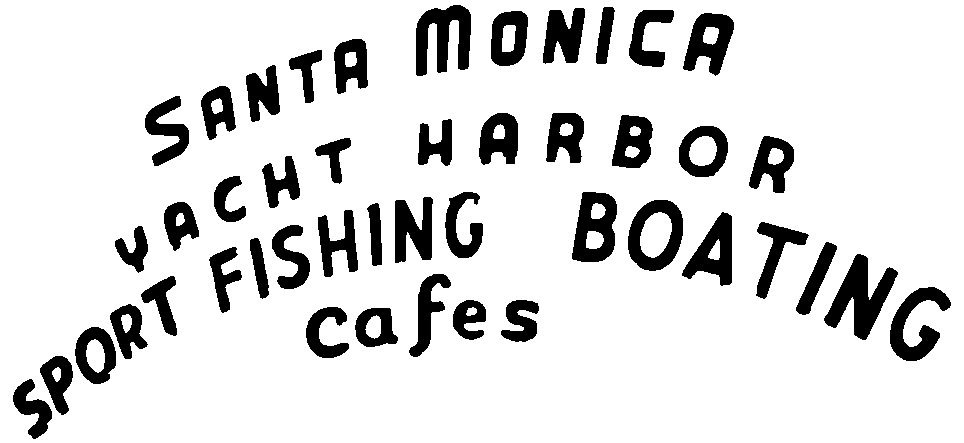}}}\end{subfigure}
		\begin{subfigure}{0.23\textwidth}{{\includegraphics[height=0.34\linewidth, width=0.9\linewidth]{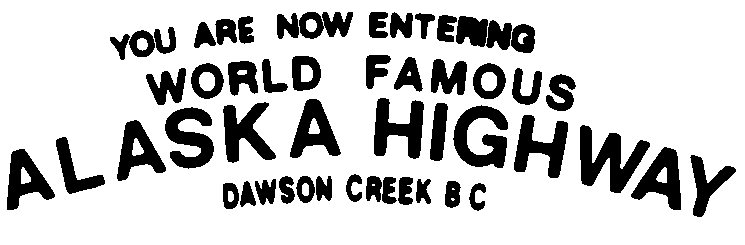}}}\end{subfigure}
		\begin{subfigure}{0.23\textwidth}{{\includegraphics[height=0.34\linewidth, width=0.9\linewidth]{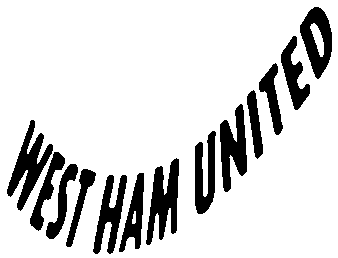}}}\end{subfigure}
		\begin{subfigure}{0.23\textwidth}{{\includegraphics[height=0.34\linewidth, width=0.9\linewidth]{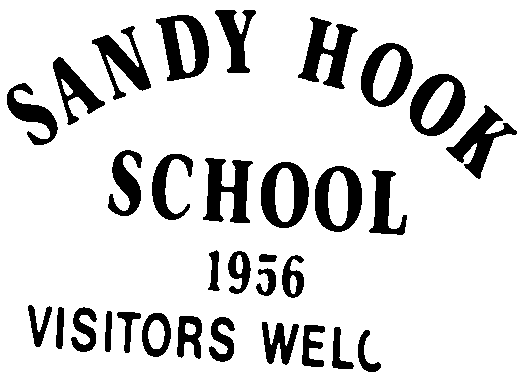}}}\end{subfigure}
	\end{center}
	\caption{Examples of pixel-level annotation (cropped) in Total-Text.}\vspace{-.1in}
	\label{fig:pixelanno}
\end{figure*}
\textbf{Evaluation Protocol.} Like ICDARs datasets \cite{survey}, Total-Text uses DetEval\cite{wolf2006object}. We did a modication to the minimum intersection area calculation stage to handle our polygon-shaped groundtruth. The evaluation protocol will be made available as well. 
%A new evaluation protocol is needed to handle the proposed groundtruth format. We built the evaluation protocol on top of Wolf and Jolion \cite{wolf2006object}. Similar to the conventional method, we modified the protocol so that it could calculate the minimum intersection area between ground truth and detected polygon. Suppose we have polygon points of $i$-th text line from the dataset ground truth $\textbf{p}_k^g=\{p_1,p_2,\ldots,p_n\}_k^g$, and the estimated polygon points $\textbf{p}_k^e=\{p_1,p_2,\ldots,p_n\}_k^e$, the minimum intersection area $a_k$ is defined as $a_k = \frac{ (\text{Area}(\textbf{p}_k^e) \cap \text{Area}(\textbf{p}_k^g)) }{(\text{Area}(\textbf{p}_k^e) \cup \text{Area}(\textbf{p}_k^g))-(\text{Area}(\textbf{p}_k^e) \cap \text{Area}(\textbf{p}_k^g))}$. 

\textbf{Annotation Details.} Groundtruth in the Total-Text is annotated in word level granularity. Adopted from the COCO-text, word level texts are \textit{uninterrupted sequence of characters separated by a space}. As mentioned, Total-Text uses polygon shapes to bind groundtruth words tightly. Apart from that, we also included rectangular bounding box annotation considering most of the current algorithms generate rectangule bounding box outputs. However, it is not an accurate representation as a big chunk of background area is included due to the nature of curved text. Therefore, we do not encourage the usage of rectangular bounding box in our dataset. Total-Text considers only English characters in natural images; other languages, digital watermarks and unreadable texts are labelled as \textit{do not care} in the groundtruth. \textit{Do not care} area picks up by algorithms should be filtered out before evaluating its performance. Groundtruth for word recognition is also provided along with its spatial coordinates. In addition, orientation of every instances were annotated for modularity convenience. For example, if one prefer to evaluate curved text detection ability only, one could leverage this annotation to filter out intances with other orientations. Last but not least, Total-Text also comes with binary mask groundtruth to cater the recent requirements \cite{he2016accurate,Zhang_2016_CVPR,yao2016scene}. Fig. \ref{fig:cover} illustrates all the aforementioned annotation details apart from the pixel-level annotation, which is illustrated in Fig. \ref{fig:pixelanno}. Considering the scale of this dataset is manageable, authors of this paper annotated the entire dataset manually and cross checked with another 3 laboratory members.

\subsection{Dataset Statistics}
This subsection will discuss the statistics of Total-Text. All of the comparisons are made against ICDAR 2013 and MSRA-TD500, as they are the most common benchmark for horizontal and multi-oriented focused scene text respectively. Total-Text is split into two groups, training and testing set with 1255 and 300 images, respectively.

\textbf{Strength in numbers.} Fig. \ref{fig:stats} shows a series of statistics information of the Total-Text. It has a total of 9330 annotated texts, 6 instances per image in average. More than half of the images in Total-Text have 2 different orientations and above, yielding 1.8 orientations per image on average. Both numbers ranked first against its competitors \cite{survey}, showing the complexity of Total-Text. Apart from these solid numbers, the dataset was also collected with quality in mind, including scene complexity such as text-like and low contrast background, different font types and sizes, etc, image examples in Fig. \ref{fig:examples_b}. 

\begin{figure*}[ht]
	\begin{center}
		\includegraphics[height=0.3\linewidth, width=0.95\linewidth]{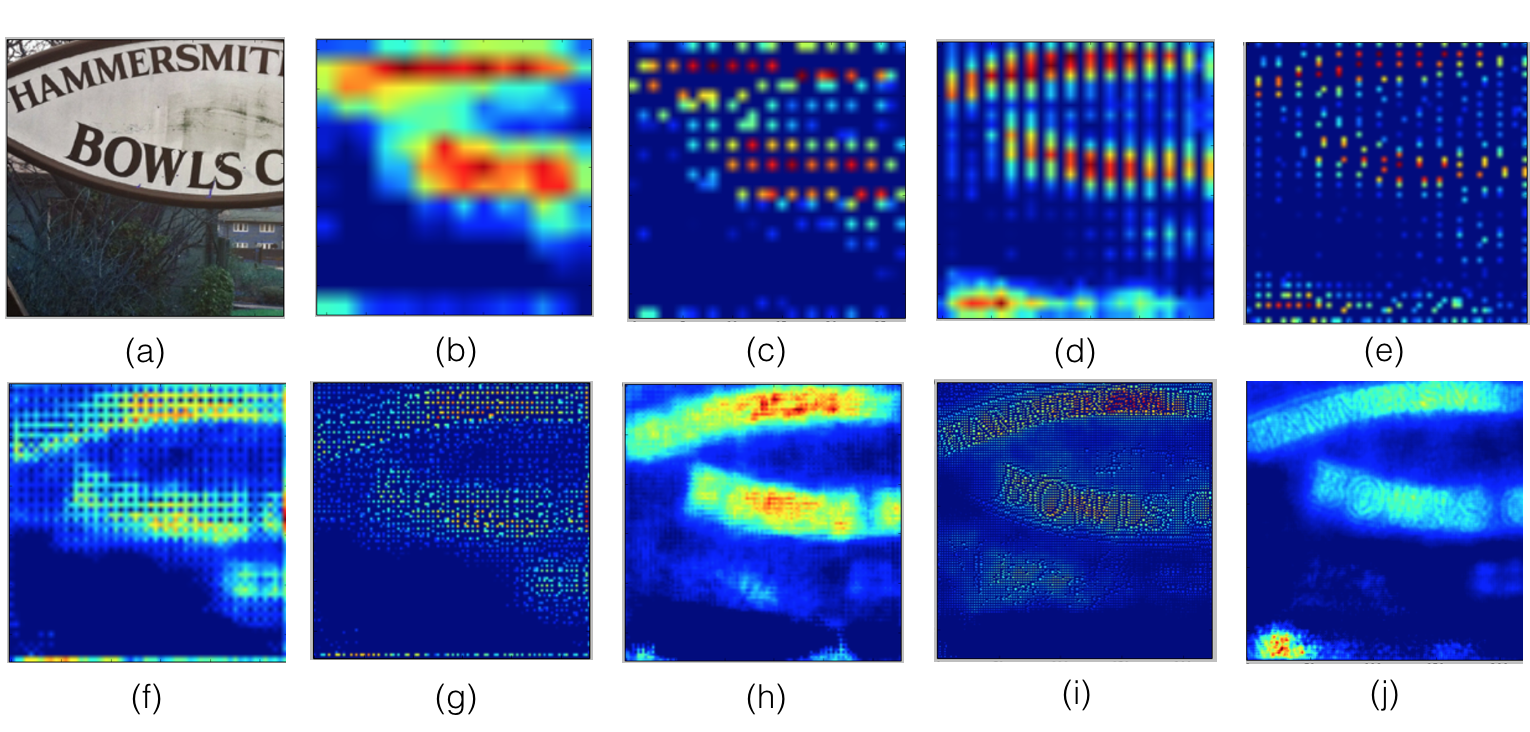}
	\end{center}
	\caption{Visualization of the activations in deconvolution network. The activation maps from top left to bottom right correspond to the output maps from lower to higher layers in the deconvolution network. We select the most representative activation in each layer for effective visualization. (a) Input image; (b) the last 14$\times$14 deconvolutional layer; (c) the 28$\times$28 unpooling layer; (d) the last 28$\times$28 deconvolutional layer; (e) the 56$\times$56 unpooling layer; (f) the last 56$\times$56 deconvolutional layer; (g) the 112$\times$112 unpooling layer; (h) the last 112$\times$112 deconvolutional layer; (i) the 224$\times$224 unpooling layer and (j) the last 224$\times$224 deconvolutional layer.} \vspace{-.1in}
	\label{fig:ctf}
\end{figure*}
\textbf{Orientation diversity.} Approximate by half of the text instances are curved, and the other half is split almost equally between horizontal and multi-oriented. Curve text has its own variation too. Based on our observation, we classified them as horizontal curved, vertical curved, circular, and wavy (refer to \ref{fig:examples_a} for image example). Their composition in the dataset can be seen in \ref{unreal11}. Although all the images were collected with curved text in mind, other orientations still occupy half of the total instances. A closer look into the dataset shows that curved text usually appears with either horizontal or multi-oriented texts. The mixture of orientations in an image, challenges text detection algorithms to achieve robustness and generalization in terms of text orientations.

\textbf{Scene diversity.} In comparison to CUTE80 (the only publicly available curved text dataset), which majority of the images are football jerseys, Total-Text is much more diversified. Fig. \ref{unreal41} shows where curved text usually appears. Business related places like restaurant (\ie~Nando’s, Starbucks), company branding logos, and merchant stores take up of 61.2\% of the curved text instances. Tourist spots such as park (\ie~Beverly Hills in America), museums and landmarks (\ie~Harajuku in Japan) occupy 21.1\%. Fig. \ref{fig:t1a} illustrates these examples.

\section{Semantic segmentation for text detection}
\label{metho}

Inspired by the success of FCN in the semantic segmentation problem, \cite{Zhang_2016_CVPR,he2016accurate,yao2016scene} casted text detection as a segmentation problem, and achieved state-of-the-art results. While most of the conventional algorithms failed in detecting curved text, their algorithms have shown successful results in limited number of examples due to the lack of available benchmark. The fact that \cite{he2016accurate} achieved good results without any heuristic grouping rules where most of the other algorithms need, intrigued us to look into this new breed of solution. We fine-tuned DeconvNet \cite{noh2015learning} and evaluated it on Total-Text, following section will discuss our findings.

\subsection{DeconvNet}

We select DeconvNet \cite{noh2015learning} as our investigation tool due to two reasons: 1) it achieved state-of-the-art results in semantic segmentation on Pascal VOC dataset and 2) Multiple deconvolutional layers in the DeconvNet allow us to observe the deviation finely. The scope of this paper is not proposing a new solution to solve the curved text problem, hence we merely convert and fine-tune the network to localize texts. For complete understanding, readers are encouraged to read \cite{noh2015learning}.

\textbf{Conversion.} The last convolution layer of the original DeconvNet has 21 layers for 20 classes in the PASCAL VOC benchmark \cite{everingham2010pascal} and one background class. In this paper, we reduced it to two layers, representing text and non-text. Then, we fine-tuned the pre-trained model provided by Noh \etal~\cite{noh2015learning} with one step training process instead of two as discussed in the original paper. Apart from these and the training data, all other training implementations were consistent with the original paper. 

\textbf{Training Data.} Considering the depth of DeconvNet (\ie~29 convolutional layers and 252M parameters), we pre-trained it using the largest scene text dataset, COCO-text \cite{veit2016cocotext}. Images in the COCO-text were categorized into legible and illegible text, where we trained our network only on the legible text as it closely resemble our dataset. Similar to \cite{he2016accurate,Zhang_2016_CVPR}, we first generated the binary mask with 1 indicating text region and 0 for background. Approximately 15k of training data were cropped into 256x256 patches to cater the receptive field of the DeconvNet. Patches with less than 10\% text regions were eliminated to prevent overwhelming amount of non-text data. Roughly 200k and 80k patches of training and validation data were generated, respectively. We augmented the data in parallel to the training with horizontal flipping and random cropping (into 224x224). %The implementation was conducted in Caffe, run on Ubuntu environment v14.05 with a single 12G TITAN-X GPU. %The training process converged after 15k iteration in 3 days' time.

\subsection{Experiments}
\textbf{Inference.} The inference process was kept to be as simple as possible. We resized input images to 224$\times$224, then forward propagated them through the DeconvNet. To generate final detection result, the saliency map was binarized using a threshold of 0.5, followed by connected component analysis to group 1s (text) pixels and bound them tightly with polygons. 

\begin{figure}[t]
	\begin{center}
		\includegraphics[height=0.5\linewidth, width=0.95\linewidth]{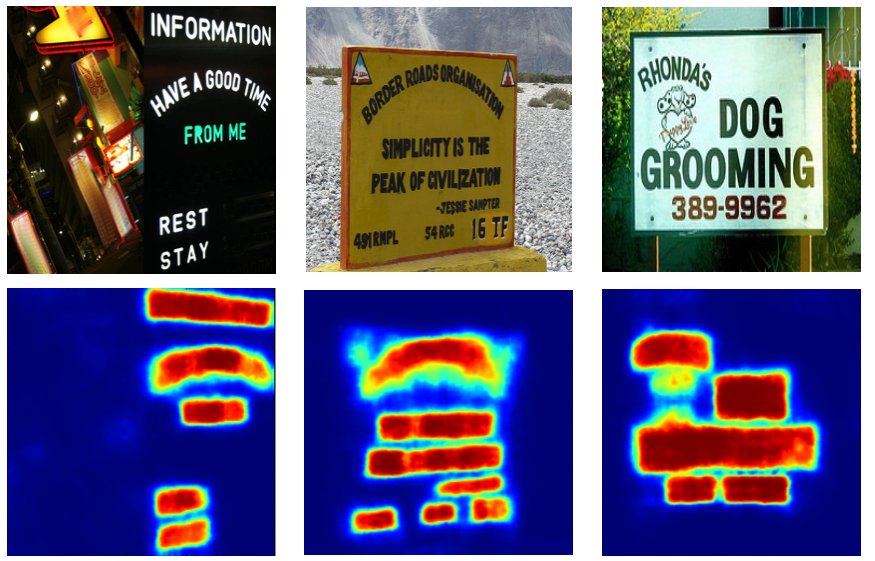}
	\end{center}
	\caption{Successful examples of DeconvNet.} \vspace{-.1in}
	\label{fig:suc}
\end{figure}

\begin{figure}[t]
	\begin{center}
		\includegraphics[height=0.5\linewidth, width=\linewidth]{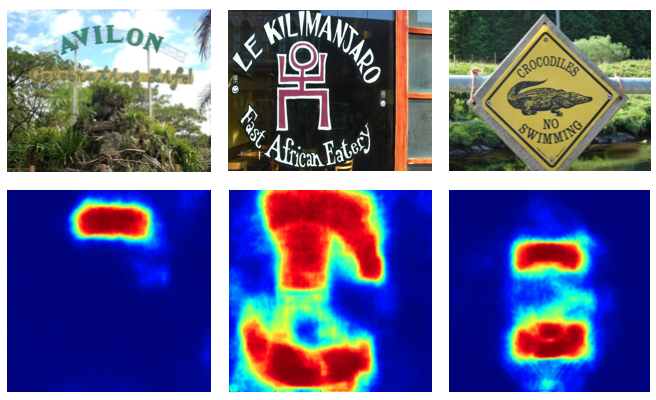}
	\end{center}
	\caption{Failure examples of DeconvNet.} \vspace{-.1in}
	\label{fig:fail}
\end{figure}

\textbf{Results.} The outcomes were evaluated using our evaluation protocol and listed in Table \ref{results}. As we went through each of the output saliency maps, we found two consistent roots that cause such unsatisfactory results: 1) The network is not robust enough for challenging backgrounds such as texts attached on repeated patterns such as bricks, gate, wall, etc.; 2) Multiple word candidates were grouped as one. Fig. \ref{fig:fail} illustrates some failure examples. We suspect the robustness of the network was affected by its training data. Such loosely bounded training data with background regions labelled as `text' could have impacted the training process to a certain extend. Meanwhile, producing word line level output is commonly seen in text detection algorithms, we lack of a segmentation process to separate them into words level.

\textbf{Deeper look into the network.} As mentioned before, our primary intention were to investigate the performance of DeconvNet on text with all sorts of orientations. With no orientation assumption or any heuristic grouping mechanism in the design, we managed to find candidates across texts with all orientations as illustrated in Fig. \ref{fig:suc}. We were curious on how and what exactly happened across the deconvolution network. So, we cropped a specific patch of an original image that consists of curved text, forward propagated through the network, and observed the feature maps in several layers of the deconvolution network. As we can see in Fig. \ref{fig:ctf}, at the lower layers, we can notice which part of the feature map is highly activated. As the layers proceed, finer details emerged, enriching the region of interest to an extend that we can recognized the characters in it.

\begin{table}[t]
	\caption{Evaluation of DeconvNet on Total-Text.}
	\label{results}
	\begin{center}
		\resizebox{8cm}{!}{
			\begin{adjustbox}{width=0.8\linewidth}
				\begin{tabular}{|l||c|c|c|}
					\hline
					Dataset & Recall & Precision & F-score \\
					\hline\hline
					Total-Text & 0.33 & 0.40 & 0.36  \\
					\hline
				\end{tabular}
		\end{adjustbox}}
	\end{center}
\end{table}

\begin{figure}[t]
	\begin{center}
		\includegraphics[height=0.5\linewidth, width=0.95\linewidth]{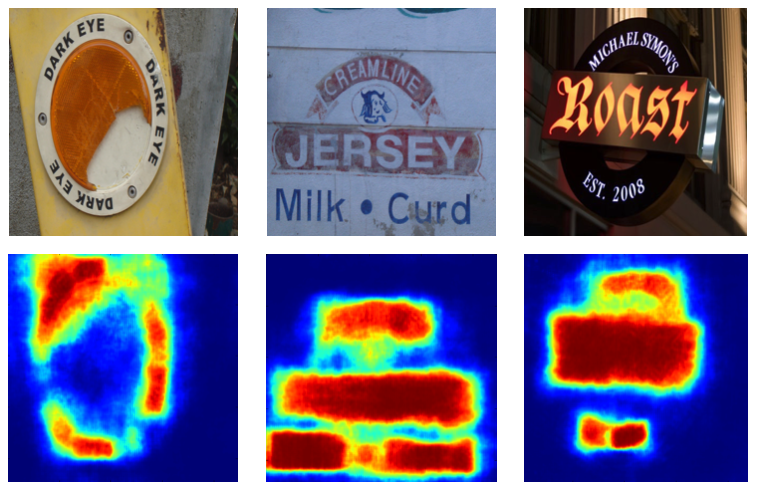}
	\end{center}
	\caption{Examples of DeconvNet with lower confidence at both end of the curved text.} \vspace{-.1in}
	\label{fig:sp}
\end{figure} 

\textbf{Spatial resolution of feature maps is crucial.} Text detection systems like \cite{Zhang_2016_CVPR, he2016accurate} adopted FCN and skip connections in their Convolutional Network. Such design element perserves spatial resolution of feature maps, and in turn provides better contextual information for their pixel-wise prediction task. Similarly, DeconvNet uses a combination of both unpooling layers and learn-able upsampling convolution filters to infer bigger feature maps layer after layer. As we can see in Fig. \ref{fig:suc}, such saliency map is high in resolution, depicts the actual shape or orientation of the detected text region. Minimal post-processing steps are required to retrieve text candidates from it. 

\textbf{Text line supervision is an interesting step forward.} Fig. \ref{fig:sp} illustrates several examples where the network is not confident about the shape of the curved text regions. We believe that it could be improved with text line supervision leveraged in \cite{he2016accurate}. This can be noticed in \cite{he2016accurate}, where the work showed their results without the FTN, its performance droped from 0.84 to 0.5 in terms of F-score.

%-----------------------------------------------
\section{Conclusion}
\label{conclude}
This paper introduces a comprehensive scene text dataset, Total-Text, featuring the missing element in current scene text datasets - curved text. We believe that curved text should be included as part of the `multi-oriented' text detection problem. While it is under research at the moment, we hope the availability of Total-Text could change the scene. We fine-tuned and analyzed how DeconvNet responds to curved text. Spatial resolution of feature maps and contextual information appeared to be crucial in segmentation based methods. Such methods are capable of predicting text regions in all sorts of orientations without hard-coded rules. Inspired by this observation, we plan to explore this area further with the aim of designing a scene text detect that is effective against multi-oriented text.

\section*{Acknowledgment}
This work is partly supported by Postgraduate Research Grant (PPP) - PG350-2016A, from University of Malaya. The Titan-X GPU used by this research was donated by NVIDIA Corporation.  We would also like to express our gratitude towards Jia Huei Tan, Yang Loong Chang and Yuen Peng Loh for Total-Text image collection and annotation. 

{\small
\bibliographystyle{IEEEtran}
\bibliography{egbib.bib}
}

\newpage
\section{Appendix}

\begin{figure*}[hb]
	\centering
	\begin{subfigure}{0.45\textwidth}
		\centering
		\includegraphics[width=\linewidth]{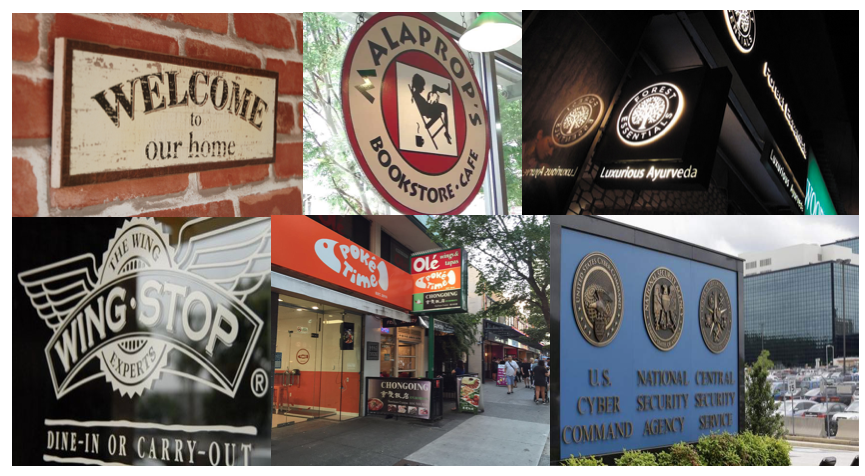}
		\caption{Perspective distorted examples.}
		\label{fig:pd}
	\end{subfigure}
	\begin{subfigure}{0.45\textwidth}
		\centering
		\includegraphics[width=\linewidth]{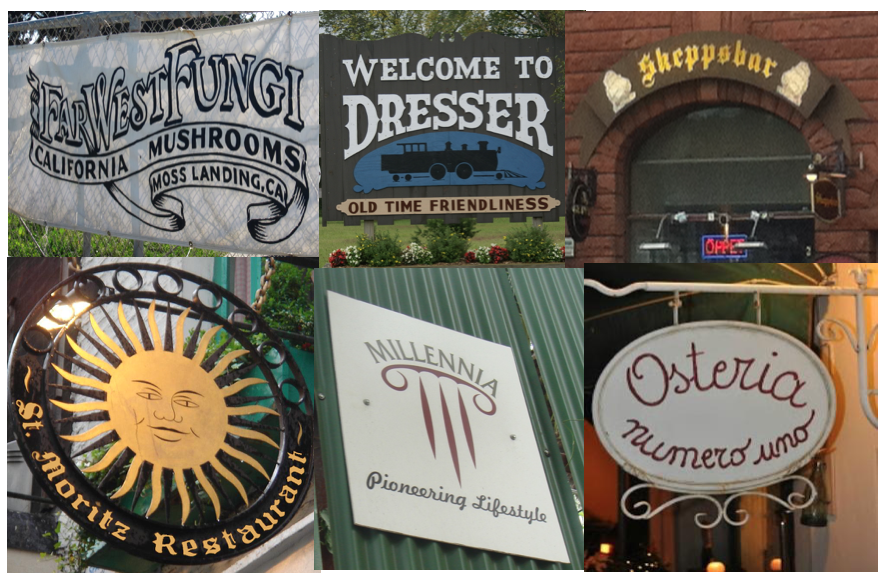}
		\caption{Different font type examples.}
		\label{fig:ft}
	\end{subfigure}
	\begin{subfigure}{0.45\textwidth}
		\centering
		\includegraphics[width=\linewidth]{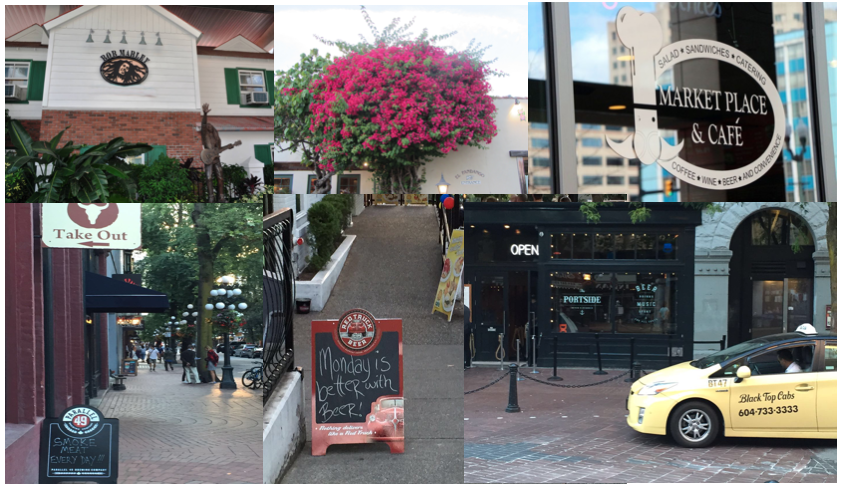}
		\caption{Different font size examples.}
		\label{fig:sf}
	\end{subfigure}
	\begin{subfigure}{0.45\textwidth}
		\centering
		\includegraphics[width=\linewidth]{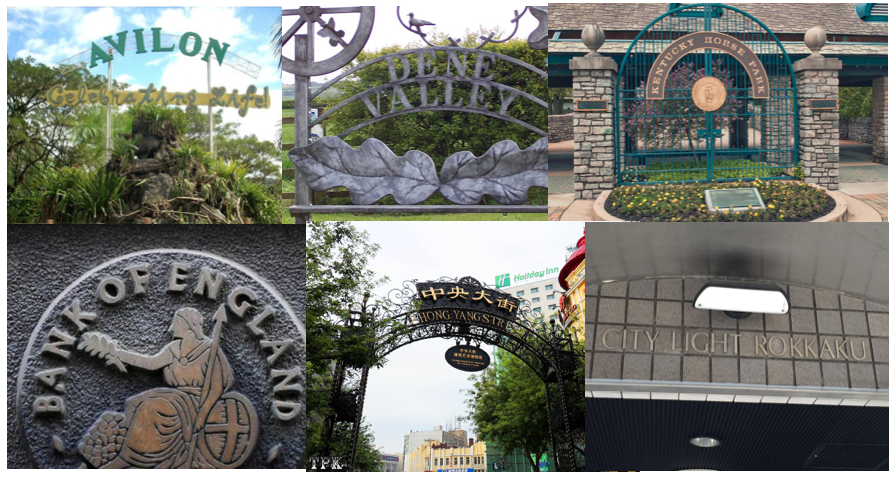}
		\caption{Complex background examples.}
		\label{fig:cb}
	\end{subfigure}
	\begin{subfigure}{0.45\textwidth}
		\centering
		\includegraphics[width=\linewidth]{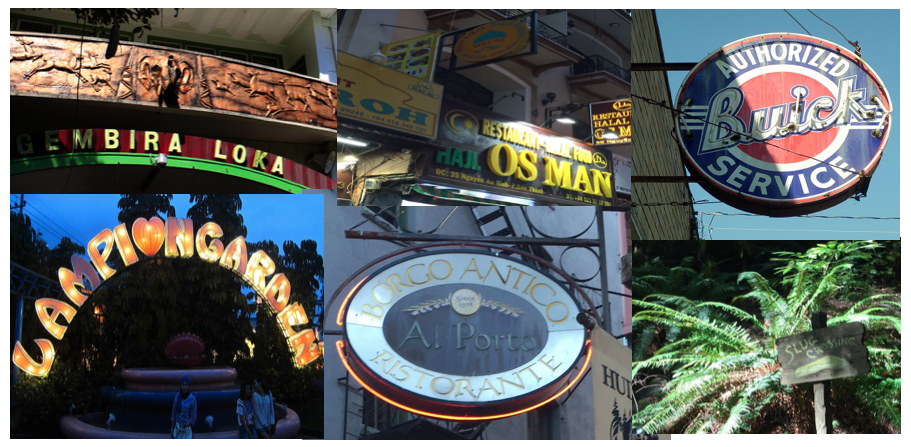}
		\caption{Uneven lighting examples.}
		\label{fig:illu}
	\end{subfigure}
	\begin{subfigure}{0.45\textwidth}
	\centering
		\includegraphics[width=\linewidth]{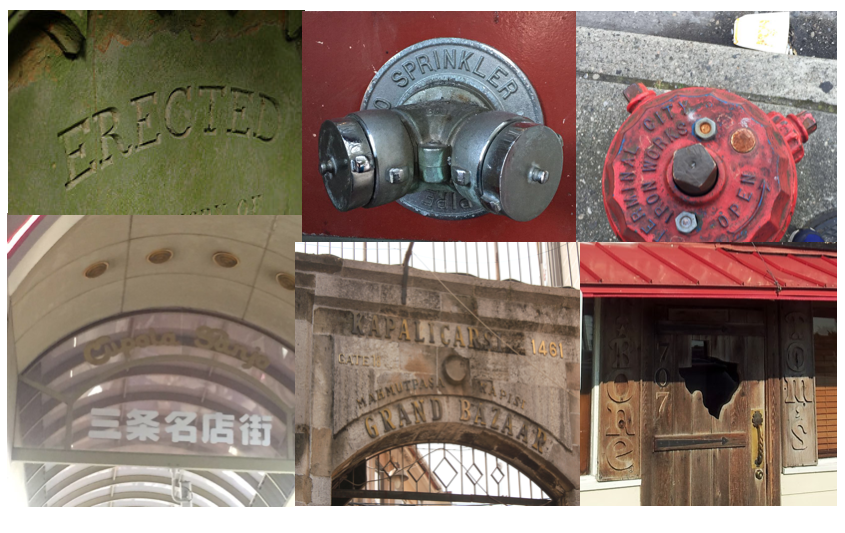}
		\caption{Low contrast examples.}
		\label{fig:lc}
	\end{subfigure}
	\caption{Challenging examples in the Total-Text dataset}
	\label{fig:realworld}
\end{figure*}

Figure \ref{fig:realworld} illustrates Total-Text dataset has very challenging attributes of real world scenery. For example, perspective distortion (Fig. \ref{fig:pd}); variation in font types (Fig. \ref {fig:ft}); variation in font sizes (Fig. \ref{fig:sf}); background with text-like characteristics such as bricks, trees etc. (Fig. \ref{fig:cb}); uneven lighting (Fig. \ref{fig:illu}) and low contrast between text and background (Fig. \ref{fig:lc}).

\subsection{Text Orientations}
\label{ornt}
\begin{figure*}[ht]
	\centering
	\begin{subfigure}{0.45\textwidth}
		\centering
		\includegraphics[width=\linewidth]{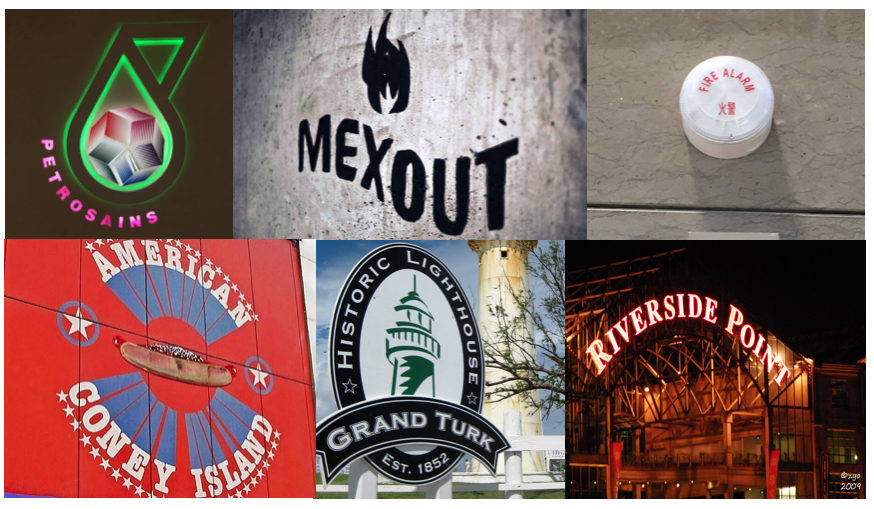}
		\caption{Curved-oriented text}
		\label{fig:c}
	\end{subfigure}
	\begin{subfigure}{0.45\textwidth}
		\centering
		\includegraphics[width=\linewidth]{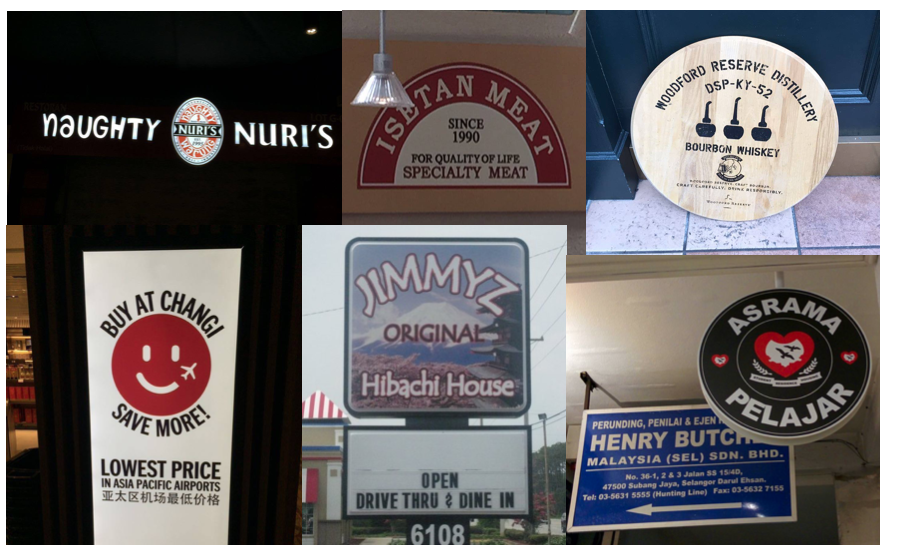}
		\caption{Curved and Horizontal-oriented text}
		\label{fig:c_h}
	\end{subfigure}
	\begin{subfigure}{0.45\textwidth}
		\centering
		\includegraphics[width=\linewidth]{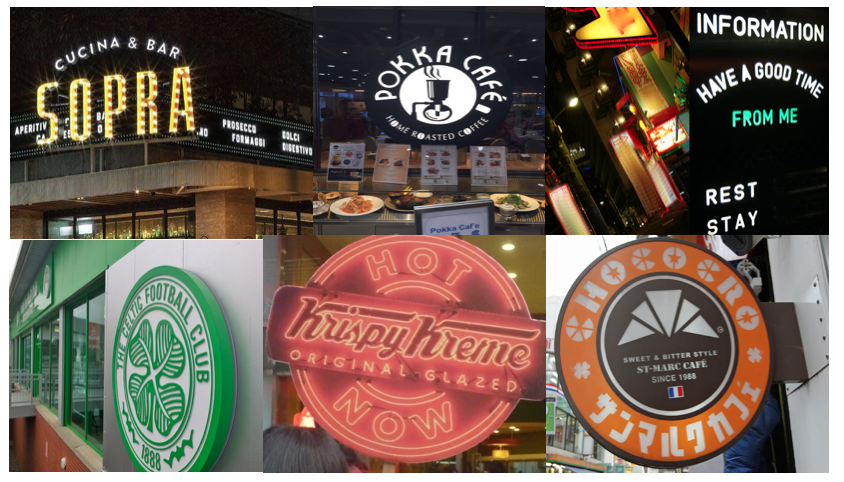}
		\caption{Curved and Multi-oriented text}
		\label{fig:c_m}
	\end{subfigure}
	\begin{subfigure}{0.45\textwidth}
		\centering
		\includegraphics[width=\linewidth]{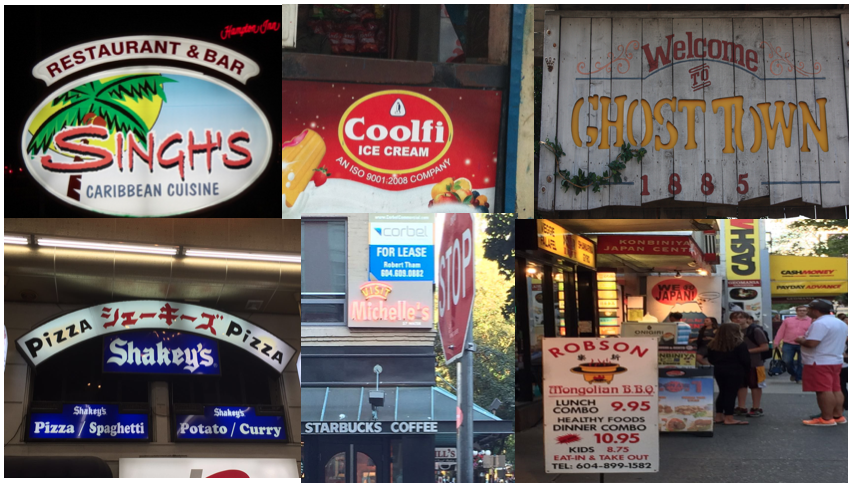}
		\caption{Curved and Horizontal and Multi-oriented text}
		\label{fig:c_h_m}
	\end{subfigure}
	\caption{Different text orientations in the Total-Text dataset}
	\label{fig:ornt}
\end{figure*}

Figure \ref{fig:ornt} shows examples with multiple text orientations. From unified orientation in Figure \ref{fig:c} to two orientations in Figure \ref{fig:c_h}-\ref{fig:c_m}, to images with all sort of orientations in Figure \ref{fig:c_h_m}.

\begin{table*}[ht]
	\begin{center}
		\resizebox{8cm}{!}{
		\begin{tabular}{|c||c |c|c||c|c|}
				\hline
				\multirow{ 2 }{*}{ Figure } & \multirow{2}{*}{Matched Ground Truth} & \multicolumn{2}{|c|}{Polygon-shaped} & \multicolumn{2}{|c|}{Rectangle-shaped}\\  \cline{3-6}  & & Precision & Recall & Precision & Recall\\
				\hline \hline
				\ref{ano1}  & 'GATE-1' & 0.27 & 1 & 0.69 & 1\\ \hline
				\multirow{2}{*}{\ref{ano2}} & 'BOULANGERLE' & 0.4 & 0.95 & 0.97 & 0.79\\ \cline{2-6} & 'patisserie' & 0.93 & 0.74 & 0.96 & 0.56\\ \hline
				\multirow{2}{*}{\ref{ano3}} & 'PURE' & 0.13 & 1 & 0.16 & 1\\ \cline{2-6} & 'ICE-CREAM' & 0.23 & 0.99 & 0.44 & 0.98\\ \hline
				\multirow{2}{*}{\ref{ano4}} & 'COSTA' & 0.86 & 0.64 & 0.95 & 0.43\\ \cline{2-6} & 'COFFEE' & 0.5 & 0.98 & 0.91 & 0.91\\ \hline
				\ref{ano5}  & 'Astro' & 0.48 & 0.98 & 0.9 & 0.93\\ \hline
				\ref{ano6}  & 'CLUB' & 0.43 & 1 & 0.73 & 0.94\\ \hline
				\multirow{2}{*}{\ref{ano7}} & 'INVEN' & 0.65 & 0.98 & 0.89 & 0.91\\ \cline{2-6} & 'IONS' & 0.88 & 0.57 & 0.94 & 0.45\\ \hline
				\multirow{3}{*}{\ref{ano8}} & 'GRANVILLE' & 0.37 & 0.98 & 0.86 & 0.93\\ \cline{2-6} & 'ISLAND' & 0.19 & 0.99 & 0.25 & 0.98\\ \cline{2-6} & 'anada' & 0.86 & 0.91 & 0.99 & 0.84\\ \hline
				\multirow{2}{*}{\ref{ano9}} & 'JEWELRY' & 0.39 & 0.99 & 0.95 & 0.73\\ \cline{2-6} & 'MARKET' & 0.18 & 1 & 0.48 & 0.945\\ \hline
		\end{tabular}}
		\caption{Evaluation results with different groundtruth format. Our proposed polygon-shaped groundtruth, provided alongside Total-Text bounds text regions tightly and hence provide a more accurate evaluation result.}
		\label{tab:diffgt}
	\end{center}
\end{table*}

\subsection{Different Groundtruth Evaluations}
\label{diffgt}

\begin{figure*}[ht]
\centering
\begin{subfigure}{0.3\textwidth}
  \centering
  \includegraphics[height=\linewidth, width=\linewidth]{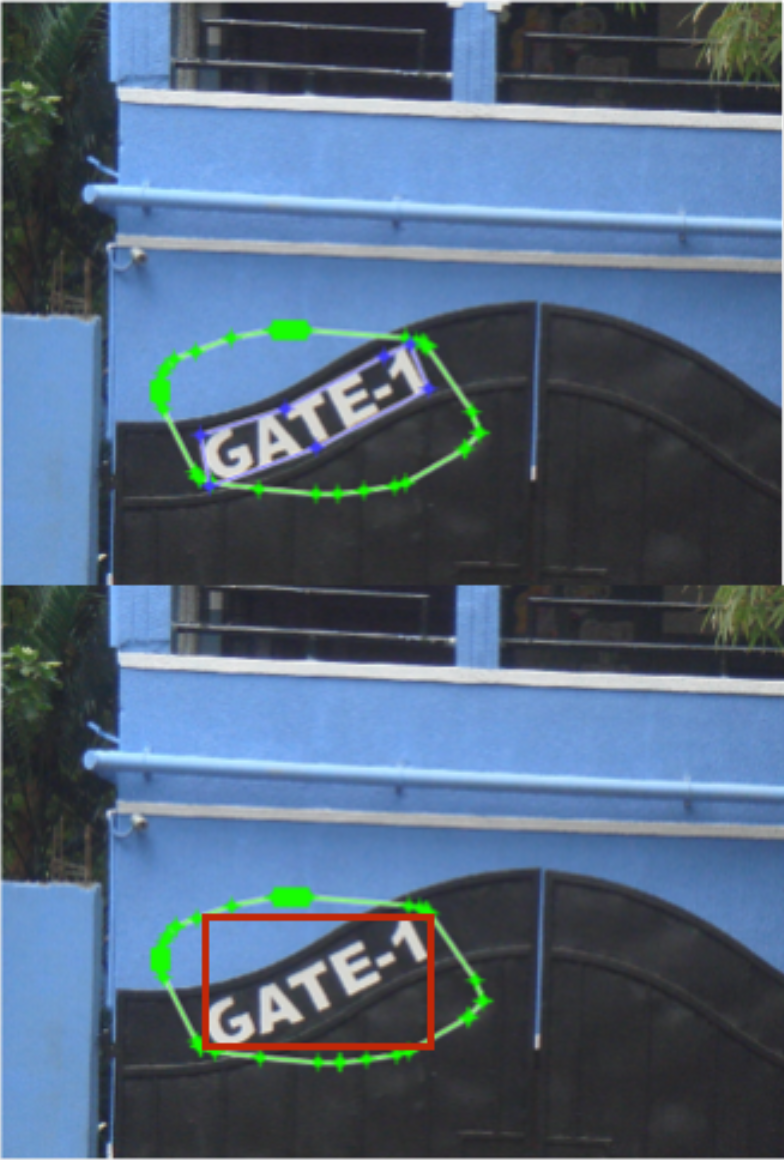}
  		\caption{}
		\label{ano1}
\end{subfigure}
\begin{subfigure}{0.3\textwidth}
  \centering
  \includegraphics[height=\linewidth, width=\linewidth]{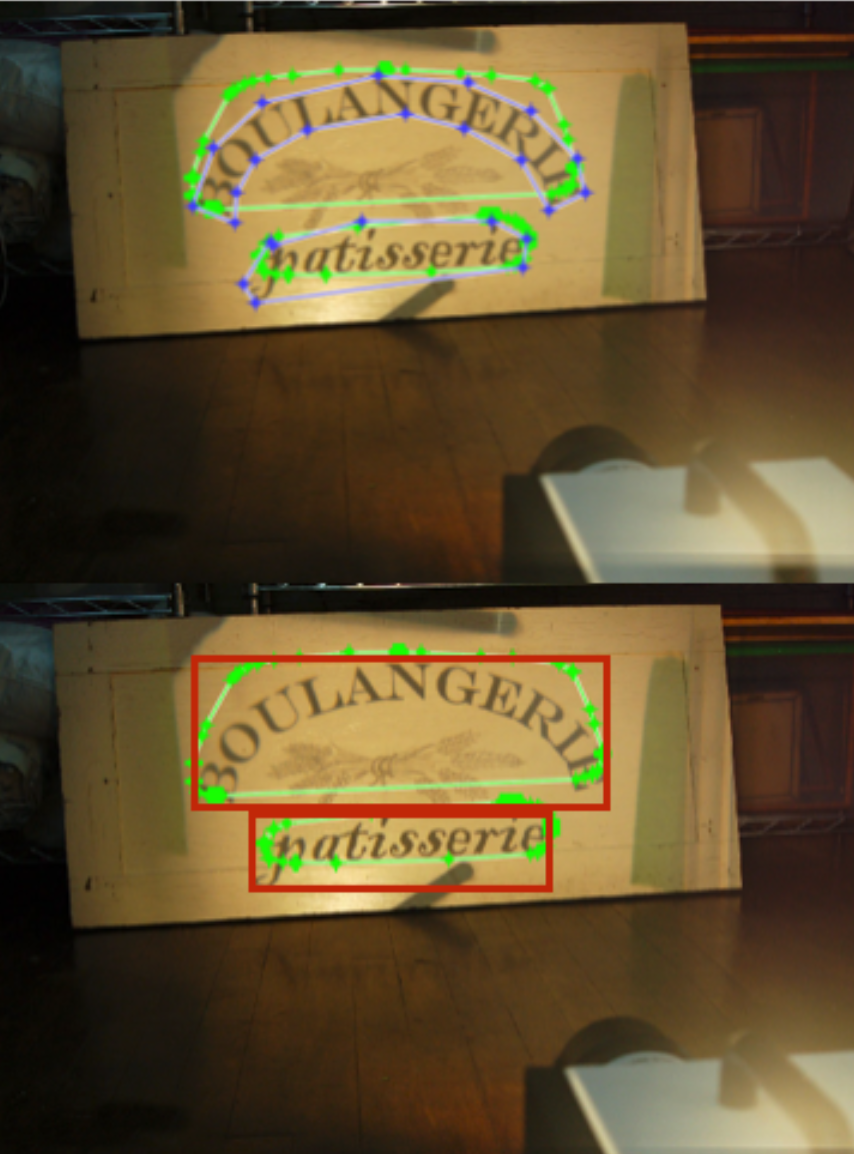}
    		\caption{}
		\label{ano2}
\end{subfigure}
\begin{subfigure}{0.3\textwidth}
  \centering
  \includegraphics[height=\linewidth, width=\linewidth]{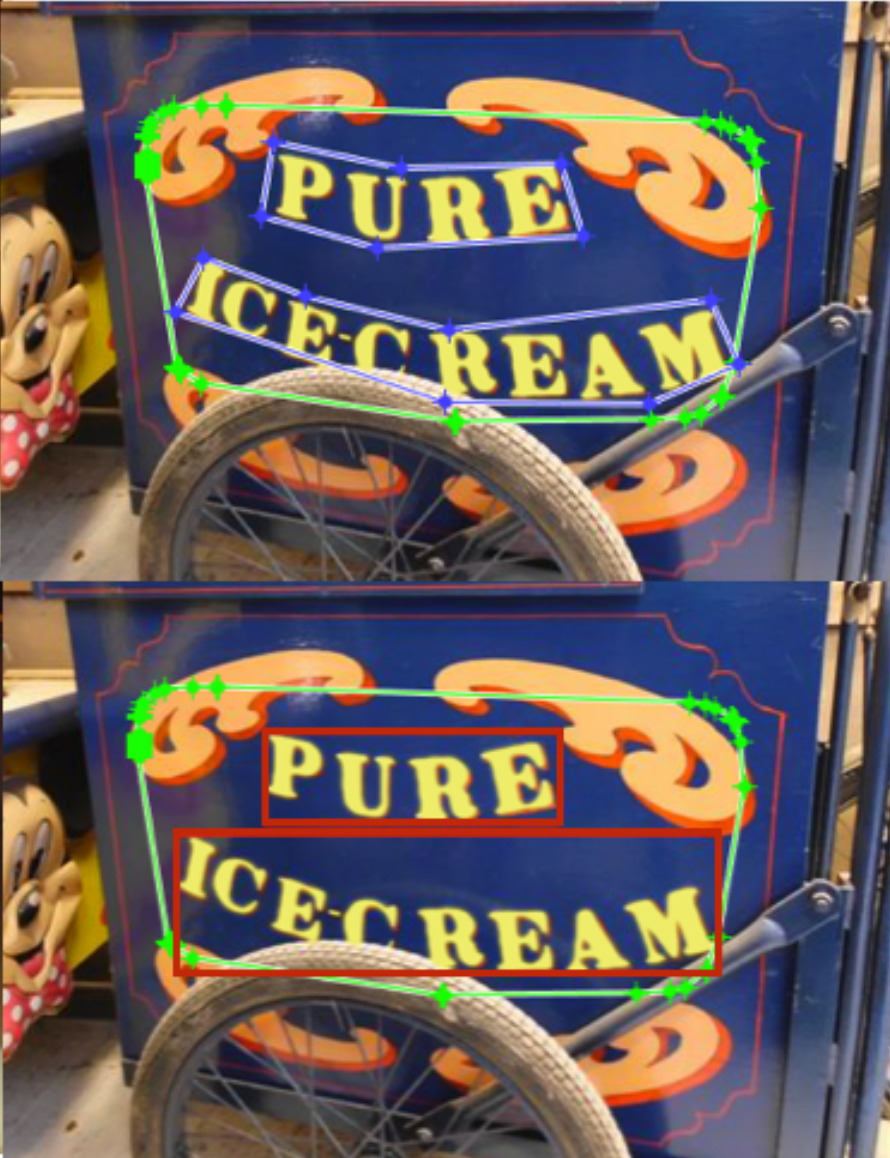}
    		\caption{}
		\label{ano3}
\end{subfigure}
\begin{subfigure}{0.3\textwidth}
  \centering
  \includegraphics[height=\linewidth, width=\linewidth]{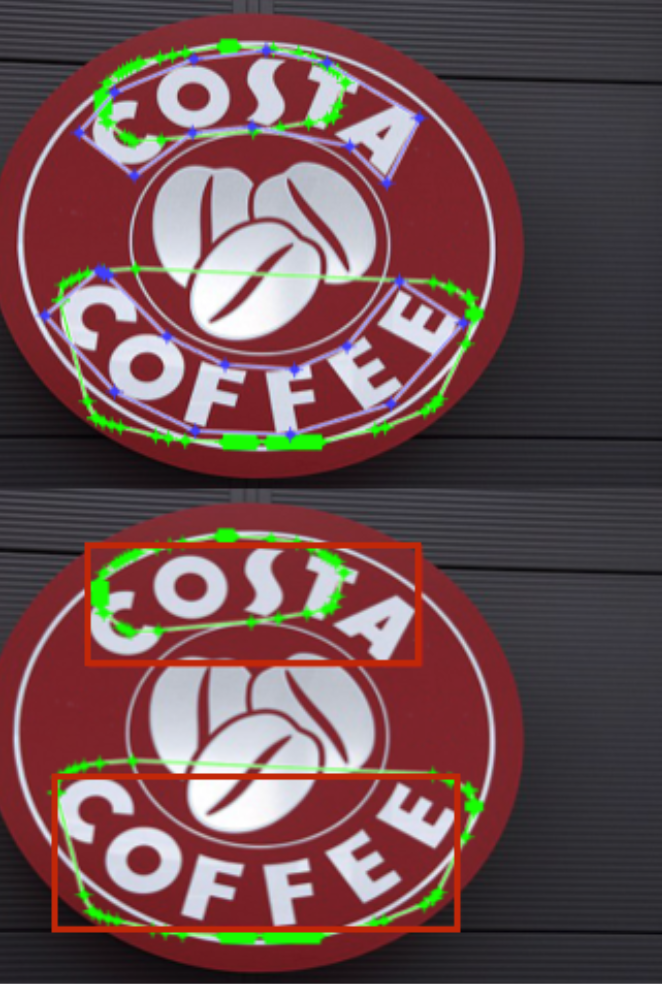}
    		\caption{}
		\label{ano4}
\end{subfigure}
\begin{subfigure}{0.3\textwidth}
  \centering
  \includegraphics[height=\linewidth, width=\linewidth]{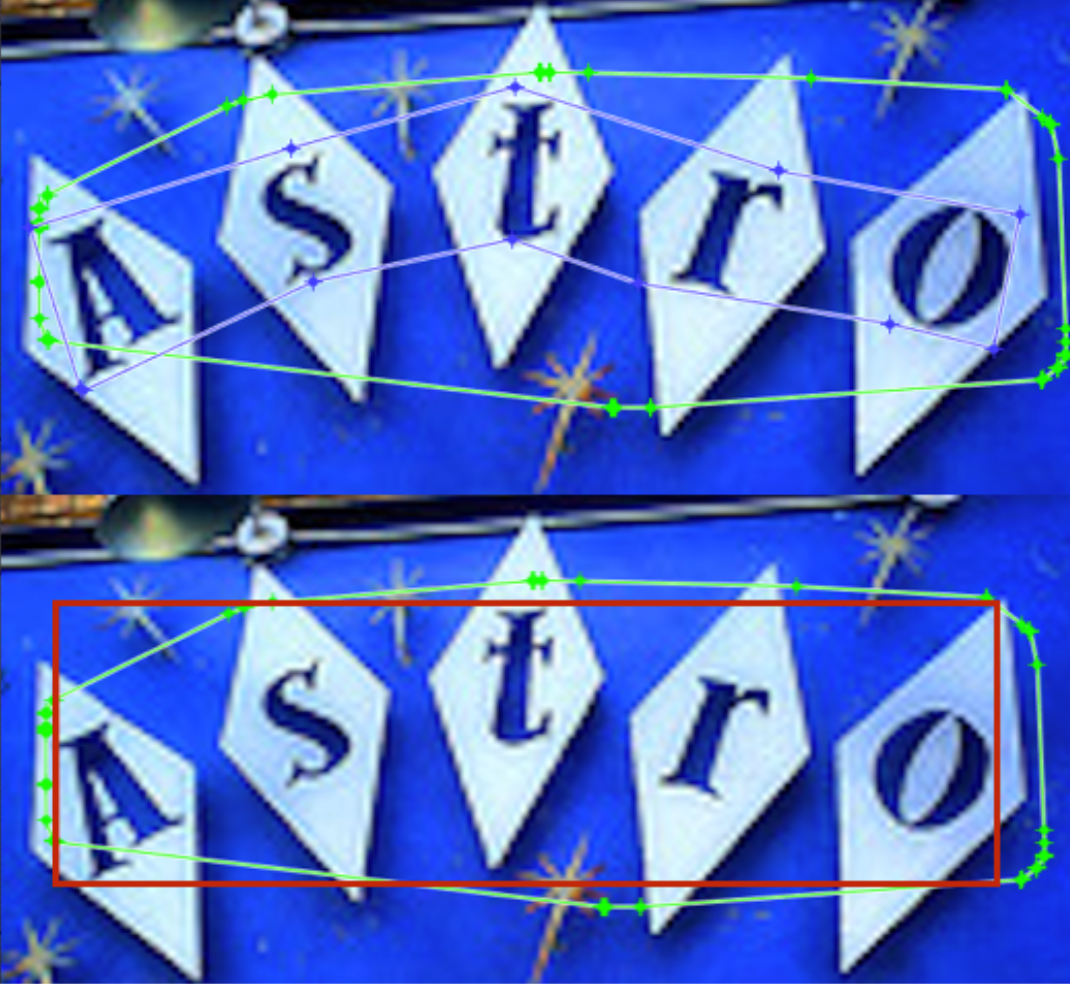}
    		\caption{}
		\label{ano5}
\end{subfigure}
\begin{subfigure}{0.3\textwidth}
  \centering
  \includegraphics[height=\linewidth, width=\linewidth]{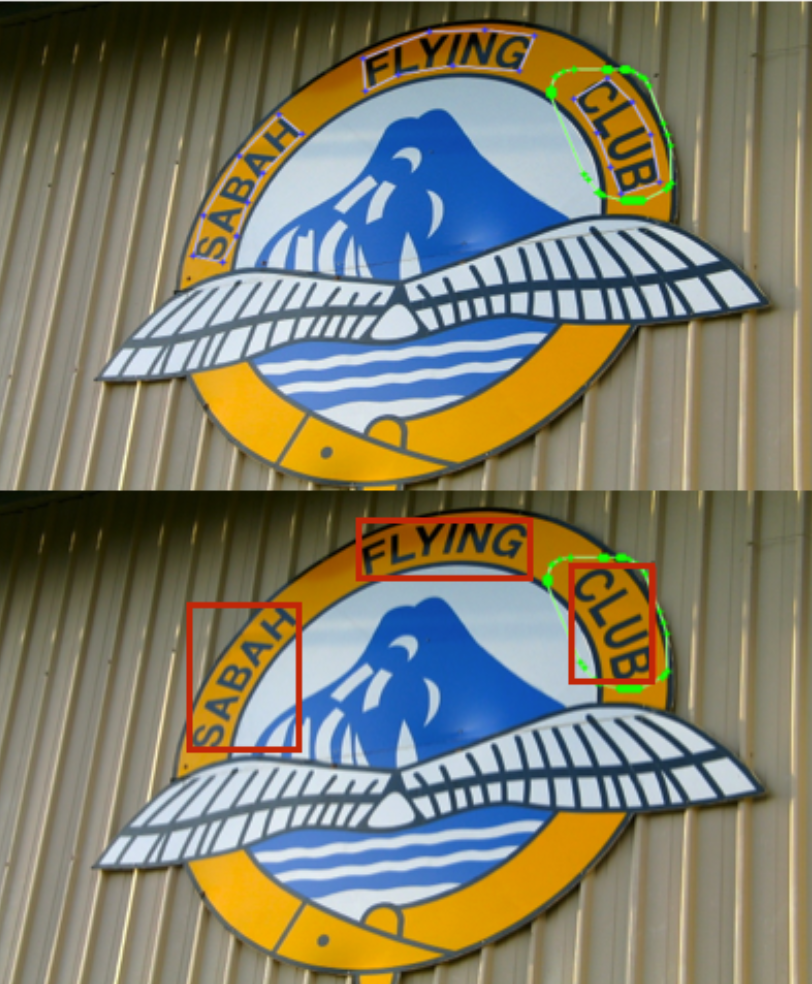}
    		\caption{}
		\label{ano6}
\end{subfigure}
\begin{subfigure}{0.3\textwidth}
  \centering
  \includegraphics[height=\linewidth, width=\linewidth]{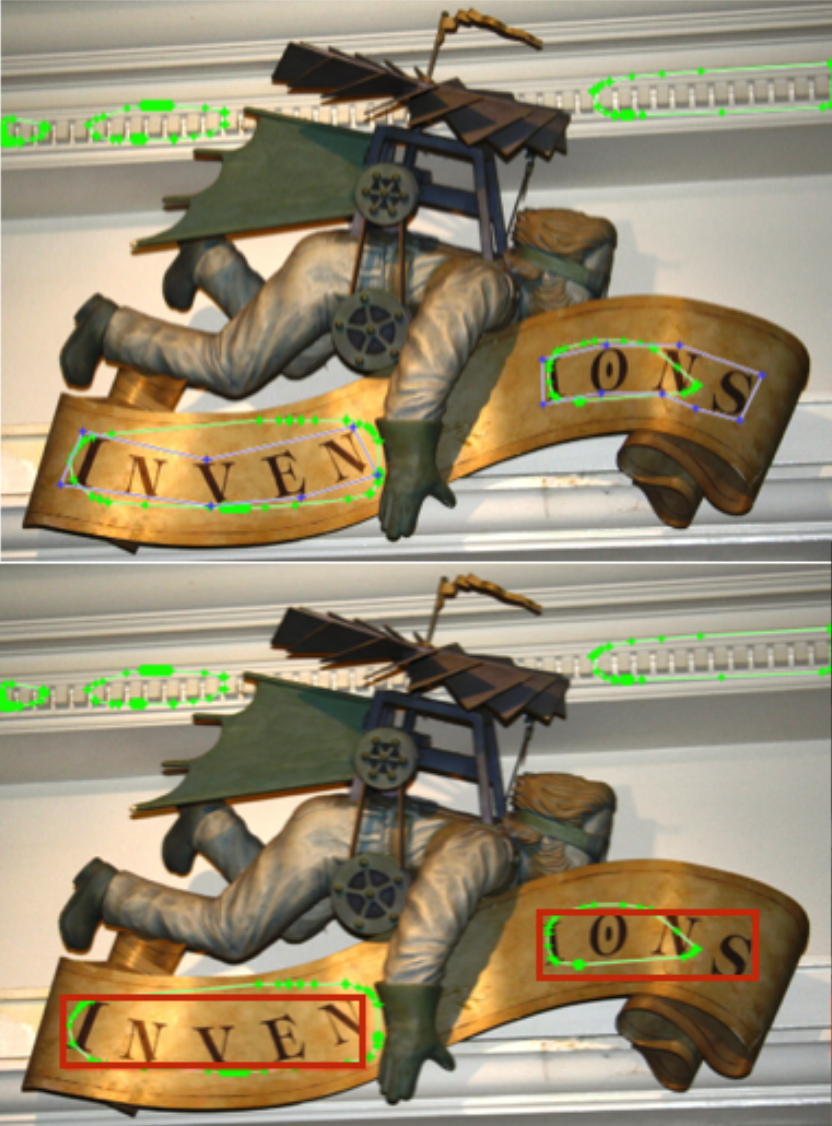}
    		\caption{}
		\label{ano7}
\end{subfigure}
\begin{subfigure}{0.3\textwidth}
  \centering
  \includegraphics[height=\linewidth, width=\linewidth]{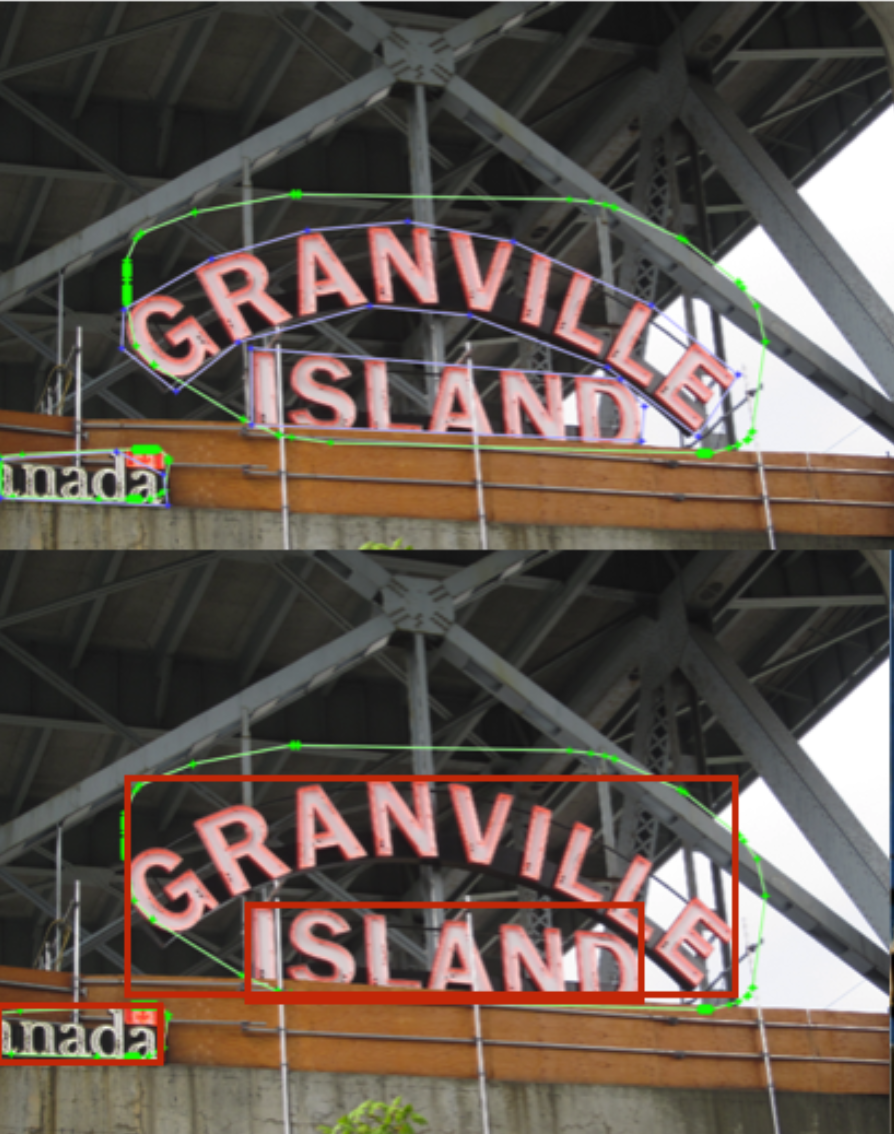}
    		\caption{}
		\label{ano8}
\end{subfigure}
\begin{subfigure}{0.3\textwidth}
  \centering
  \includegraphics[height=\linewidth, width=\linewidth]{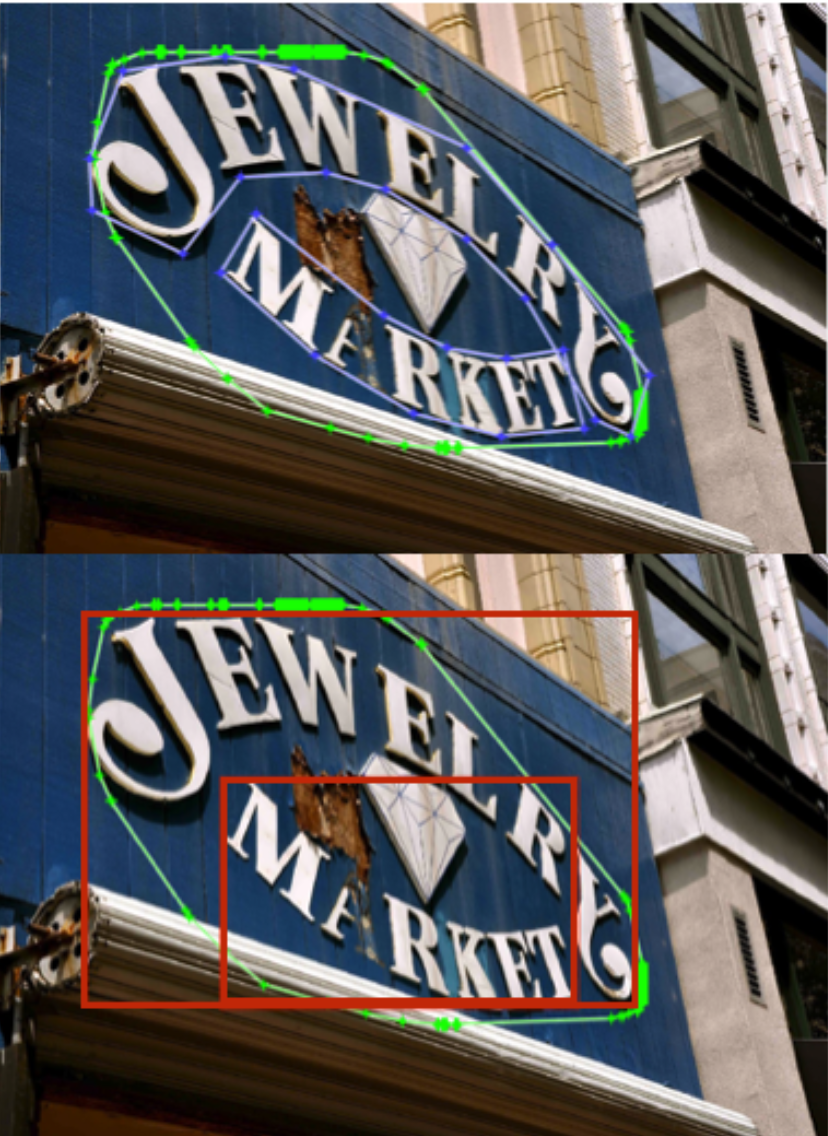}
    		\caption{}
		\label{ano9}
\end{subfigure}
\caption{Disagreement between polygon-shaped (in {\color{blue}blue} colour) and rectangle-shaped (in {\color{red}red} colour) groundtruth regions. It is found out that evaluation results using the polygon-shaped groundtruth provides a more accurate representation of algorithm's performance.}
\label{fig:diffgt}
\end{figure*}

Table \ref{tab:diffgt} and Figure \ref{fig:diffgt} show the comparison between two different groundtruth in terms of {\it Precison} and {\it Recall}. Note that, {\color{green}Green} is the detected text using the DeconvNet algorithm \cite{noh2015learning}, {\color{red}Red} is the groundtruth generated using the conventional rectangular box and {\color{blue}Blue} is the groundtruth generated using our proposed polygon-shape. Figure \ref{ano1}-\ref{ano3} show examples of the detected text (in {\color{green}green} region) has higher precision score if we choose to employ the conventional rectangle-shaped groundtruth ({\color{red}red} in color). Figure \ref{ano4}-\ref{ano9} illustrate several examples of the detected text (in {\color{green}green} region) have lower recall and precision because it misses a large intersection area with the groundtruth regions.

\subsection{Groundtruth Examples}

\begin{figure*}[ht]
\centering
\begin{subfigure}{0.45\textwidth}
  \centering
	\includegraphics[height=\linewidth, width=\linewidth]{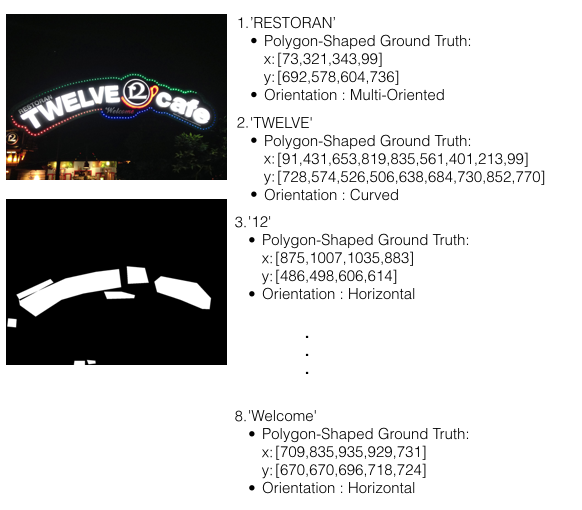}
  \label{fig:1}
\end{subfigure}
\quad
\begin{subfigure}{0.45\textwidth}
  \centering
	 \includegraphics[height=\linewidth, width=\linewidth]{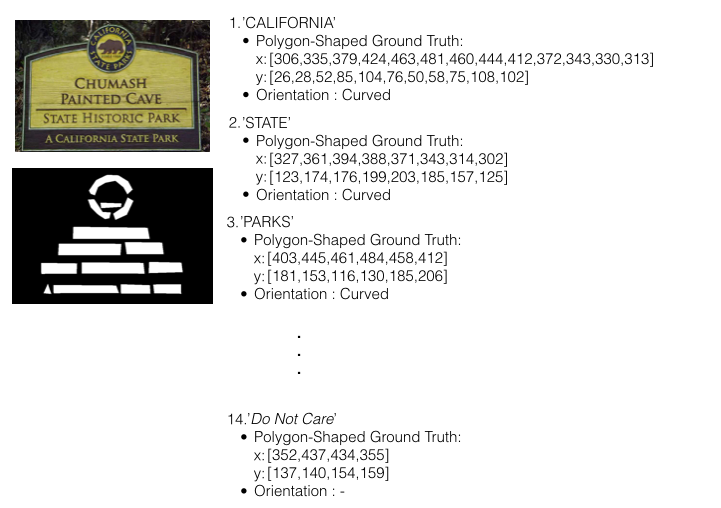}
   \label{fig:2}
\end{subfigure}
\quad
\begin{subfigure}{0.45\textwidth}
  \centering
	  \includegraphics[height=\linewidth, width=\linewidth]{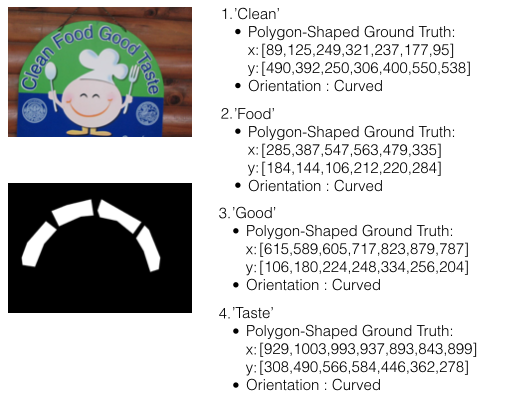}
  \label{fig:3}
\end{subfigure}
\begin{subfigure}{0.45\textwidth}
  \centering
	  \includegraphics[height=\linewidth, width=\linewidth]{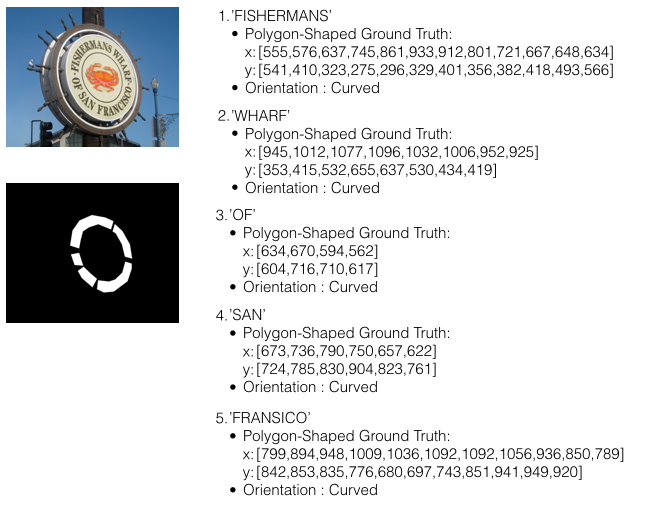}
  \label{fig:4}
\end{subfigure}
\caption{Examples of Total-Text annotations with polygon-shaped groundtruth. It can be noticed that the polygon-shaped bounding box tightly bounded the text, the annotations are more comprehensive.}
\label{fig:exm1}
\end{figure*}

\begin{figure*}[ht]
\centering
\begin{subfigure}{0.45\textwidth}
  \centering
  \includegraphics[height=\linewidth, width=\linewidth]{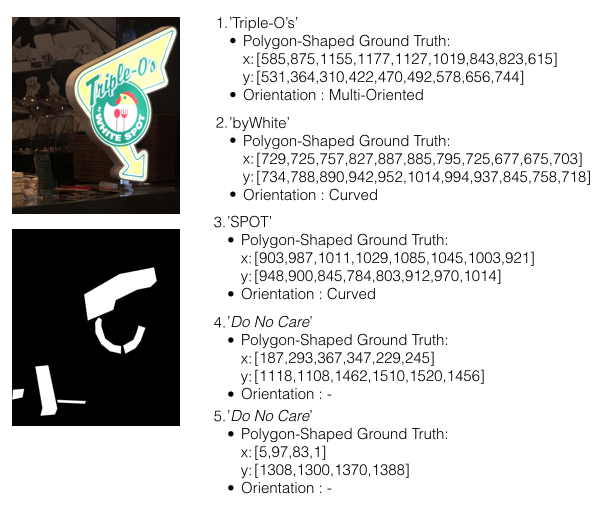}
 \label{fig:7}
\end{subfigure}
\begin{subfigure}{0.45\textwidth}
  \centering
  \includegraphics[height=\linewidth, width=\linewidth]{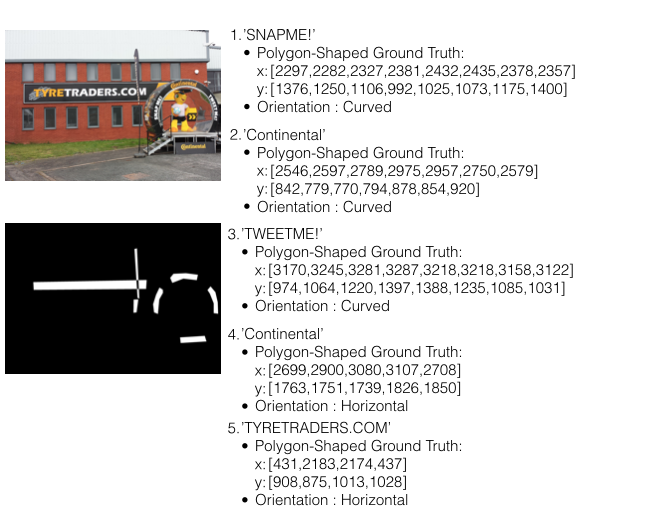}
   \label{fig:8}
\end{subfigure}
\begin{subfigure}{0.45\textwidth}
  \centering
  \includegraphics[height=\linewidth, width=\linewidth]{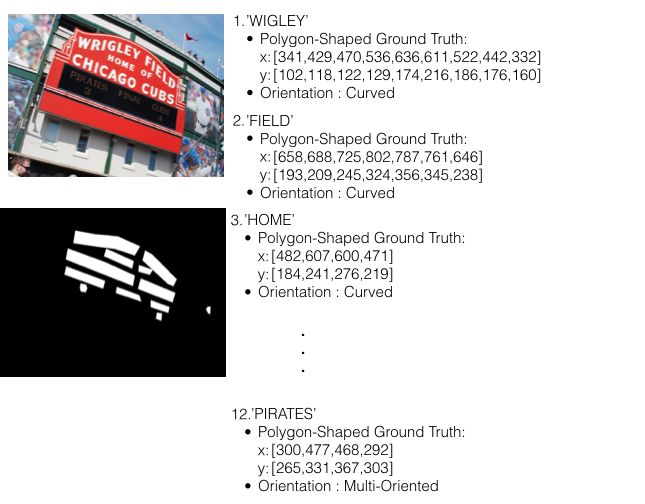}
 \label{fig:9}
\end{subfigure}
\begin{subfigure}{0.45\textwidth}
  \centering
  \includegraphics[height=\linewidth, width=\linewidth]{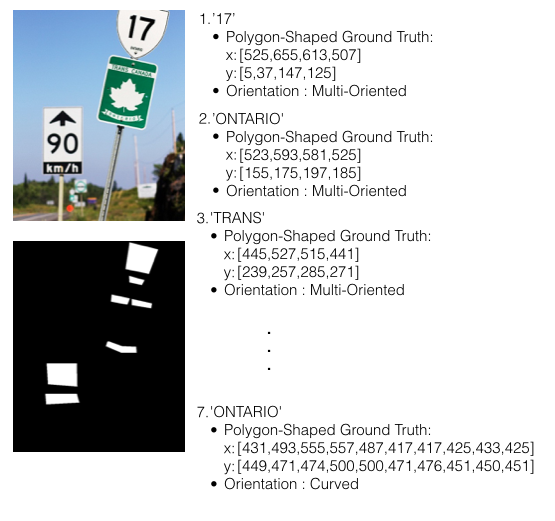}
  \label{fig:10}
\end{subfigure}
\caption{More examples of Total-Text annotations with polygon-shaped groundtruth.}
\label{fig:exm2}
\end{figure*}

Figure \ref{fig:exm1} and \ref{fig:exm2} depict the annotation details of Total-Text dataset. Every images were annotated into four attributes: 1) spatial locations, 2) transcript, 3) orientation of each text instances, and 4) binary mask with annotated region as 1(white), background as 0(black).

% that's all folks
\end{document}